\documentclass{article}

 \usepackage[main, final]{neurips_2026}

\usepackage[utf8]{inputenc} 
\usepackage[T1]{fontenc}    
\usepackage{hyperref}       
\usepackage{url}            
\usepackage{booktabs}       
\usepackage{amsfonts}       
\usepackage{nicefrac}       
\usepackage{microtype}      
\usepackage{xcolor}         
\definecolor{mygreen}{RGB}{34, 139, 34}
\definecolor{myblue}{RGB}{70, 130, 180}
\definecolor{mypink}{RGB}{255, 105, 180}
\definecolor{myorange}{RGB}{244, 164, 96}
\usepackage{graphicx}
\usepackage{amsmath}
\title{AVI-HT: Adaptive Vision-IMU Fusion \\ for 3D Gloved Hand Tracking}

\author{%
Ziyi Kou \quad Ankit Kumar \quad Mia Huang \quad Taylor Niehues \\
\textbf{Vatsal Mehta \quad Ergys Ristani \quad Li Guan} \\[4pt]
Meta Reality Labs
}

\begin{document}

\maketitle

\begin{abstract}

We present \textbf{AVI-HT}, an adaptive visual-IMU fusion approach for tracking 3D hand poses by jointly modeling the egocentric images with on-glove 6-DoF IMU signals. AVI-HT significantly improved accuracy and availability, particularly in hand-object interaction (HOI) scenarios involving heavy visual occlusion. Two complementary ingredients underpin its success: (1) synchronized multi-modal training data pairing on-body vision-IMU sensor streams with ground-truth 3D hand poses from a motion-capture system, and (2) a cross-sensor deep attention mechanism that adaptively modulates the trust assigned to the vision and individual IMU sensors. To evaluate AVI-HT in real-world settings, we conduct extensive experiments on our DexGloveHOI dataset that consists of 100K+ pairwise vision-IMU samples with synchronized 3D annotated poses, in which users manipulate a variety of objects during daily tasks. We compare against multiple single- and multi-modal tracking approaches under two hand models (UmeTrack, MANO). The results show that AVI-HT reduces mean keypoint error by 16.1\% and its wrist-aligned variant by 24.2\% over the baselines. 
A hand-object contact evaluation shows that AVI-HT keeps finger contacts with the manipulated object more accurately and consistently over time, reducing contact losses by 8.5\%.
Ablation studies further show that our fusion design outperforms simpler schemes and that the gains grow with occlusion and hold across subjects.
\end{abstract}

\begin{figure}[t]
\centering
\includegraphics[width=0.9\linewidth]{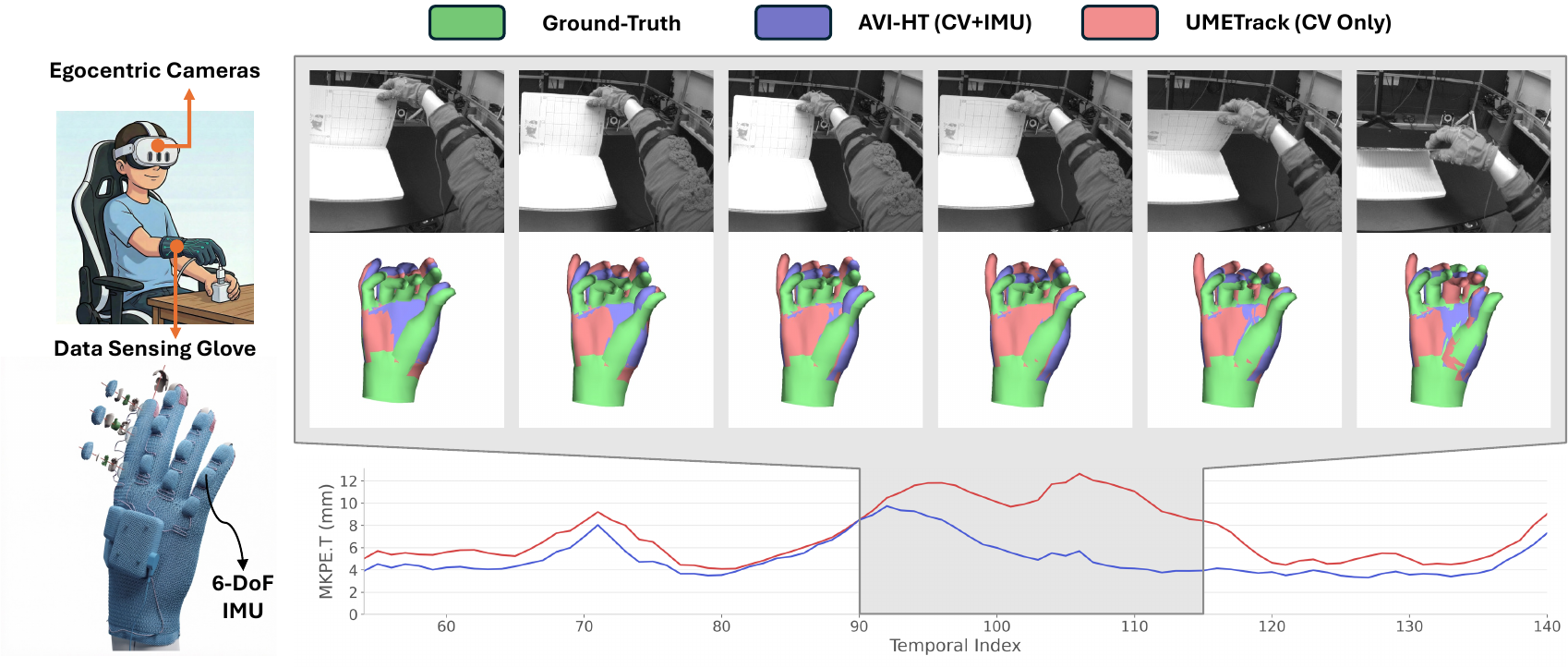}
    \caption{\textbf{3D hand tracking from egocentric vision and on-glove IMUs.} \emph{(Left)} Our multi-modal data capture setup that consists of an egocentric camera and a data sensing glove with 12 6-DoF IMU sensors. \emph{(Right)} Qualitative comparison on 4 egocentric frames. For each frame, the top row shows an inset hand diagram. The bottom row shows the corresponding 3D hand pose tracking overlaid for \textcolor{mypink}{UMETrack} and \textcolor{myblue}{AVI-HT} against \textcolor{mygreen}{ground truth}. The tracking results by AVI-HT are closely aligned with the ground truth, while the vision-only baseline deviates substantially. The top yellow patches highlight the activated IMU sensors by the adaptive attention module of AVI-HT, indicating the regions with greater reliance on IMU signals over egocentric vision. \emph{Best viewed in color.}}
    \label{fig:intro}
\end{figure}

\section{Introduction}

Consider the set of egocentric images in the first row of Figure \ref{fig:intro}, where a human hand manipulates the note book on the table, which is common in AR/VR \cite{zhang2020mediapipe,moon2020interhand2} and robotic teleoperation \cite{qin2023anyteleop,arunachalam2023dexterous} applications. As the grasp tightens, the fingers are progressively occluded. These interactions are contact-rich and happening in 3D, yet the very act of grasping hides the hand from the camera. 
Therefore, accurately estimating hand pose from egocentric vision alone becomes extremely difficult under occlusion, as the set of kinematically plausible configurations that explain the visible evidence could be prohibitively large \cite{zimmermann2017learning}.
To mitigate such limitations, we introduce a framework with a data sensing glove that can accurately perceive hand pose in 3D from joint visual and on-glove IMU inputs, particularly when the hand is heavily occluded by the object it is manipulating.

Recent developments in computer vision and robotics point to the direction where advances are achieved by leveraging joint visual and sensor inputs, powered by multi-modal fusion models. This emerging insight has been demonstrated in the context of computer vision by ImageBind \cite{girdhar2023imagebind}, MultiMAE \cite{bachmann2022multimae}. In the context of wearable sensing and human motion capture, we observe this with systems like \cite{pan2023fusing} and \cite{bao2022fusepose}. In the area of 3D hand pose tracking, prior work has largely relied on a single modality: either RGB images \cite{pavlakos2024reconstructing,rong2021frankmocap} or inertial sensors \cite{sarker2026real}. Each of these approaches suffers from fundamental limitations when applied to dexterous hand-object interaction (HOI) scenarios, such as inaccurate hand reconstruction due to heavy occlusion from vision signals \cite{mueller2017real}, magnetic interference from 9-DoF IMUs \cite{maereg2017low}, or lack of absolute orientation from 6-DoF IMUs \cite{sarker2026real}. 

In this paper, we take the philosophy of multi-modal fusion and apply it to the problem of 3D hand pose tracking. 
\emph{Our motivation is that visual signals and 6-DoF IMUs naturally complement each other: vision provides absolute position and global orientation from wrist but degrades in finger tracking under occlusion, while 6-DoF IMUs capture high-frequency local joint dynamics that are invariant to visual obstruction but lack a global  spatial reference. Therefore, fusing the two recovers what each modality alone cannot.}
With the above motivation, we propose AVI-HT, an \underline{\textbf{a}}daptive \underline{\textbf{v}}isual-\underline{\textbf{I}}MU fusion approach for 3D \underline{\textbf{h}}and \underline{\textbf{t}}racking from joint egocentric images and on-glove 6-DoF IMU signals. AVI-HT captures faithful 3D hand poses in a variety of dexterous interaction scenarios, especially when evaluated on challenging HOI scenarios with heavy occlusion.
As illustrated in the second row of Figure \ref{fig:intro}, a vision-only approach UMETrack \cite{han2020megatrack} produces hand pose estimates that deviate substantially from the ground truth under heavy occlusion, whereas our joint vision-IMU approach recovers poses that are more closely aligned.

The key to the success of AVI-HT lies in two complementary innovations. Firstly, for training data, we build a synchronized multi-modal capture setup by integrating egocentric cameras with a data sensing glove, and deploy a motion capture (MoCap) system for 3D ground-truth pose collection. This yields a dataset of paired egocentric video and per-finger IMU measurements with accurate 3D annotations. Secondly, for the model, we design a deep cross-sensor attention-based architecture that adaptively fuses visual features with on-glove IMU signals. As visualized in Figure \ref{fig:intro}, there are 2 to 3 IMU sensors mounted at each finger of the glove. When a finger joint is occluded, indicated in egocentric images, AVI-HT automatically places greater reliance on the IMU signal at that location to compensate for the loss of visual evidence.
Conversely, the wrist remains largely visible to the egocentric cameras even during heavy object manipulation, which provides a consistent estimate of global wrist rotation and translation that anchors the entire hand in 3D space.
The combination of two ingredients leads to significant improvements compared to vision-only and IMU-only works.

Evaluating multi-modal hand tracking methods in realistic conditions remains difficult: existing datasets are largely captured in controlled, low-occlusion settings and rarely include inertial sensor data \cite{zimmermann2019freihand,banerjee2025hot3d}. To enable rigorous evaluation under the dexterous manipulation scenarios that matter most, we collect DexGloveHOI, a multi-subject evaluation dataset focusing on hand-object-interaction (HOI). The dataset pairs egocentric video with time-synchronized on-glove IMU signals and marker-based 3D ground-truth hand poses across a diverse range of activities. Using DexGloveHOI, we evaluate AVI-HT with two hand representations, UMETrack \cite{han2022umetrack} and MANO \cite{MANO:SIGGRAPHASIA:2017}, and conduct extensive experiments comparing AVI-HT with single- and multi-modal approaches. Results demonstrate that IMU signals provide complementary cues that yield more stable and accurate hand tracking performance, with the largest gains in heavily occluded regions.


In summary, we contribute AVI-HT, an approach for 3D hand tracking from egocentric images and on-glove IMU signals. We demonstrate the key effect of cross-sensor attention for handling occlusion and achieve state-of-the-art results on DexGloveHOI. Beyond keypoint accuracy, AVI-HT keeps hand-object contacts more accurately and consistently over time. Extensive ablations show that our fusion design outperforms simpler schemes, the gains grow with occlusion and hold across subjects, and each IMU sensor contributes most to its own finger, providing practical insights for deploying multi-modal hand tracking in AR/VR and robotic teleoperation.

\section{Related Work}

\textbf{3D Hand Pose Tracking.} In recent years, significant progress has been made in estimating 3D hand pose and shape from a single RGB image \cite{lin2021two, guo20223d}. Earlier methods often relied on convolutional neural networks to regress parameters of parametric hand models \cite{boukhayma20193d} or directly estimate 3D joint locations \cite{rong2021frankmocap}. More recently, transformer-based architectures have emerged as the dominant paradigm. Methods such as METRO \cite{lin2021end} and MeshGraphormer \cite{lin2021mesh} leverage self-attention mechanisms to jointly model vertex-vertex and vertex-joint interactions for end-to-end hand mesh reconstruction. Building on this, HaMeR \cite{pavlakos2024reconstructing} demonstrated that scaling up both the training data and the vision transformer capacity leads to substantial improvements for in-the-wild hand reconstruction. However, while these vision-based methods use large-scale data to improve generalization, they fundamentally suffer from severe performance degradation under heavy occlusion, which is a common occurrence during dexterous hand-object interactions \cite{li2025handnet}. On the other hand, wearable inertial measurement unit (IMU) sensors offer an occlusion-free alternative. 
Existing IMU-based hand tracking systems typically rely on dense arrays of 9-DoF sensors \cite{mummadi2018real}. However, the magnetometer in 9-DoF IMUs is highly susceptible to electromagnetic interference in indoor environments \cite{laidig2023vqf}. While more compact 6-DoF IMUs avoid magnetometer-related distortions, they lack of an absolute orientation reference and suffer from cumulative gyroscope drift \cite{sarker2026real}. 
In this paper, we propose AVI-HT, an adaptive visual-IMU fusion framework that overcomes the occlusion limitations of vision-only methods while leveraging sparse, low-cost 6-DoF IMUs for high-fidelity global hand pose reconstruction.

\textbf{Visual-Sensor Fusion for Pose Estimation.} Fusing visual data with inertial sensor signals has proven highly effective for resolving ambiguities in human poses. For full-body tracking, methods like DIP-IMU \cite{huang2018deep} and TransPose \cite{yi2021transpose} have successfully utilized sparse IMU configurations, while subsequent works have combined these inertial priors with multi-view or egocentric cameras to achieve drift-free, occlusion-robust body tracking \cite{zhang2020fusing, li2023ego}. Despite its success, visual-sensor fusion remains a relatively unexplored domain for fine-grained 3D hand tracking. This gap is primarily due to the lack of robust, unobtrusive IMU-equipped data gloves suitable for natural manipulation, as well as the absence of high-quality multi-modal datasets that provide synchronized egocentric vision, dense IMU signals, and ground-truth 3D hand poses during complex object interactions \cite{zhang2026glove2hand}. In this paper, we address this gap by leveraging a data sensing glove covered by per-finger IMU sensors, and DexGloveHOI, a comprehensive vision-IMU dataset for multi-modal evaluation. We demonstrate how our cross-sensor attention mechanism adaptively fuses egocentric visual features with on-glove IMU signals to achieve state-of-the-art hand tracking under heavy occlusion.
\begin{figure}[t]
    \centering

\includegraphics[width=0.9\linewidth]{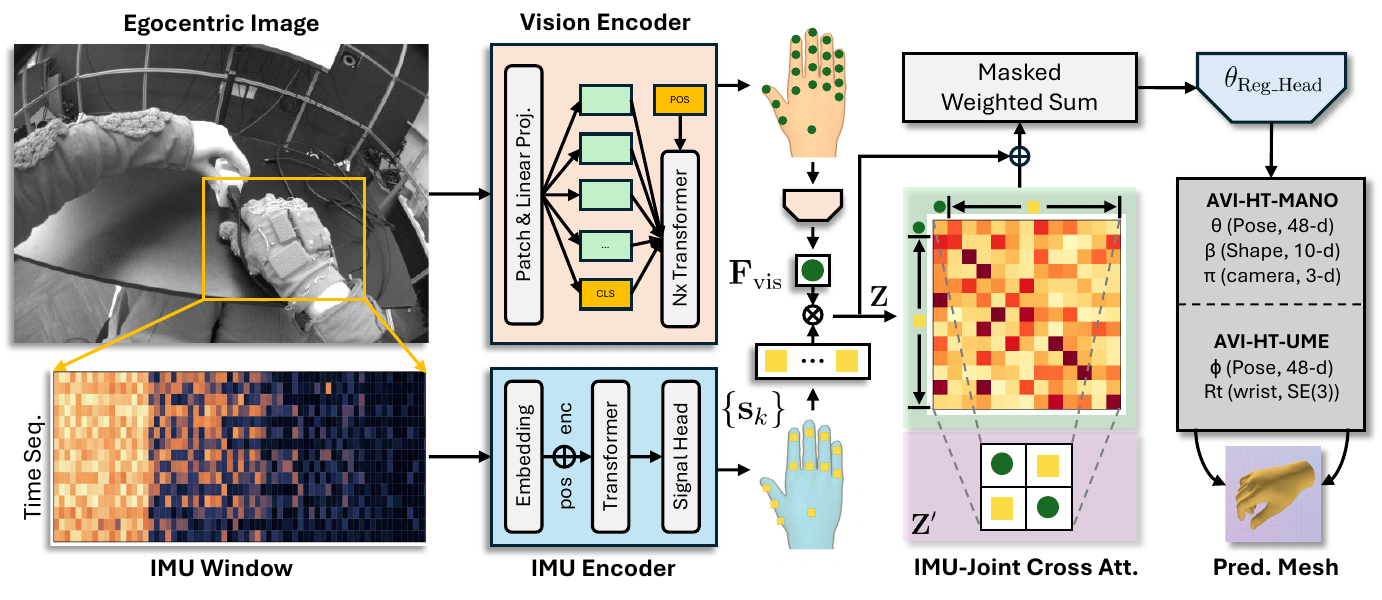}
    \caption{\textbf{Overview of AVI-HT.} A \emph{Vision Encoder} converts the egocentric image into a global visual token $\mathbf{F}_{\mathrm{vis}}$. An \emph{IMU Encoder} embeds a 14 timestamp temporal window from 12 on-glove sensors into per-sensor tokens $\{\mathbf{s}_k\}$ via a transformer encoder and signal head. The two token sets are fused through a \emph{hierarchical cross-sensor attention} module. A \emph{kinematic prior mask} aggregates attended features based on physical sensor-joint proximity while the attention scores averaged for each IMU sensor represent its adaptive importance. The fused representation is decoded by a regression head into either \textbf{MANO} or \textbf{UMETrack} parameters for hand tracking.}
    \label{fig:architecture}
\end{figure}

\section{Technical Approach}
We describe AVI-HT, our approach for 3D hand tracking from egocentric images and on-glove IMU signals. 
We show the overview of AVI-HT in Figure \ref{fig:architecture} and illustrate each component below.
\subsection{Hand Model Representations}
\label{sec:hand_models}
We evaluate AVI-HT with two hand representations to cover both offline 3D
reconstruction and real-time AR/VR tracking scenarios.
\textbf{MANO} \cite{MANO:SIGGRAPHASIA:2017} is a parametric model widely used
in offline hand mesh recovery \cite{pavlakos2024reconstructing,lin2021end,
lin2021mesh}. It defines a differentiable function $\mathcal{M}(\theta, \beta)$
that maps pose $\theta \in \mathbb{R}^{48}$ and shape $\beta \in \mathbb{R}^{10}$
to a mesh $M \in \mathbb{R}^{778 \times 3}$ and $21$ joint locations.
\textbf{UMETrack}~\cite{han2022umetrack} is an articulated model designed
for real-time egocentric hand tracking in AR/VR~\cite{han2020megatrack,
ohkawa2023efficient}. It represents pose through $22$ scalar joint angles
$\phi \in \mathbb{R}^{22}$, i.e., flexion/extension and abduction/adduction per
joint, and recovers 21 landmark positions via forward kinematics with linear blend skinning \cite{lewis2023pose}.

\subsection{Input Signals}
\label{input_signals}
\textbf{Egocentric image.} We use a Meta Quest headset that contains a monochrome egocentric camera. The camera captures egocentric images at 60\,Hz with 512x640 raw image resolution.

\textbf{On-glove IMU signals.} We use a full-hand covered data sensing glove equipped with 12 6-DoF IMU sensors. Each sensor consists of 3-axis accelerometer and 3-axis gyroscope.
The sensors are arranged across the back of the glove with one sensor on each proximal and distal phalanx of the four fingers,
with one additional sensor on the thumb and hand back each. All IMU sensors sample at 200\,Hz. Since the visual and IMU modalities operate at different rates, we synchronize them by aligning each 60\,Hz video frame to the nearest IMU timestamp. For
each aligned frame, we follow Sarker~et~al. \cite{sarker2026real} to extract a temporal window of 14 consecutive IMU samples before the synchronized timestamp. For each sample, we calculate gravity vector from 6-DoF data and append it with 3-axis gyroscope, which yields $14\times 23\times 3$ data window for each visual sample.

\textbf{3D Mocap system.} We collect 3D hand pose from a marker-based MoCap system as ground-truth. The marker positions are firstly solved to UMETrack hand pose and further converted to MANO using a Levenberg–Marquardt optimizer \cite{lourakis2005levenberg}. More description about the system is in Appendix \ref{mocap}.

\subsection{Model Architecture}
\label{sec:architecture}
AVI-HT is a general vision-IMU fusion framework that can be applied to different vision backbones and hand representations. We instantiate it with two models:  \textbf{AVI-HT-UME} for real-time AR/VR tracking and \textbf{AVI-HT-MANO} for offline 3D reconstruction. Both models share the same IMU encoder and cross-attention fusion design, differing only in the vision encoder and output head. 

\textbf{IMU encoder.} The IMU encoder is a transformer-based network \cite{vaswani2017attention} that processes the $14 \times 23\times 3$ IMU input window. It firstly applies linear mapping from $3$ to $69$ dimensions, then sinusoidal positional encoding over the 14 timesteps, followed by a 2-layer transformer encoder with multi-head self-attention over each IMU feature. The output at the last timestep is selected and projected through a two-layer MLP to produce an IMU feature vector $\mathbf{F}_{\mathrm{imu}} \in \mathbb{R}^{23\times d}$.

\textbf{Vision encoder.} \textbf{AVI-HT-UME} uses a ResNet-based encoder~\cite{he2016deep} that processes $96 \times 96$ monochrome egocentric images from two similar camera
views. A feature transform layer (FTL) \cite{han2022umetrack} lifts the 2D feature maps into a camera-disentangled 3D representation by applying SE(3) transformations derived from camera intrinsics and extrinsics. A convolutional RNN then aggregates temporal context across frames.
\textbf{AVI-HT-MANO} uses a ViT-Huge \cite{dosovitskiy2020image} backbone pre-trained on body and hand pose datasets \cite{xu2022vitpose}, which encodes a $256 \times 192$ RGB crop into a sequence of patch tokens
$\mathbf{F}_{\mathrm{vis}} \in \mathbb{R}^{N \times 1280}$. A
transformer decoder with 6 cross-attention layers then attends from a learnable query token to the patch tokens to regress MANO parameters. 

\begin{figure}[t]
    \centering

\includegraphics[width=\linewidth]{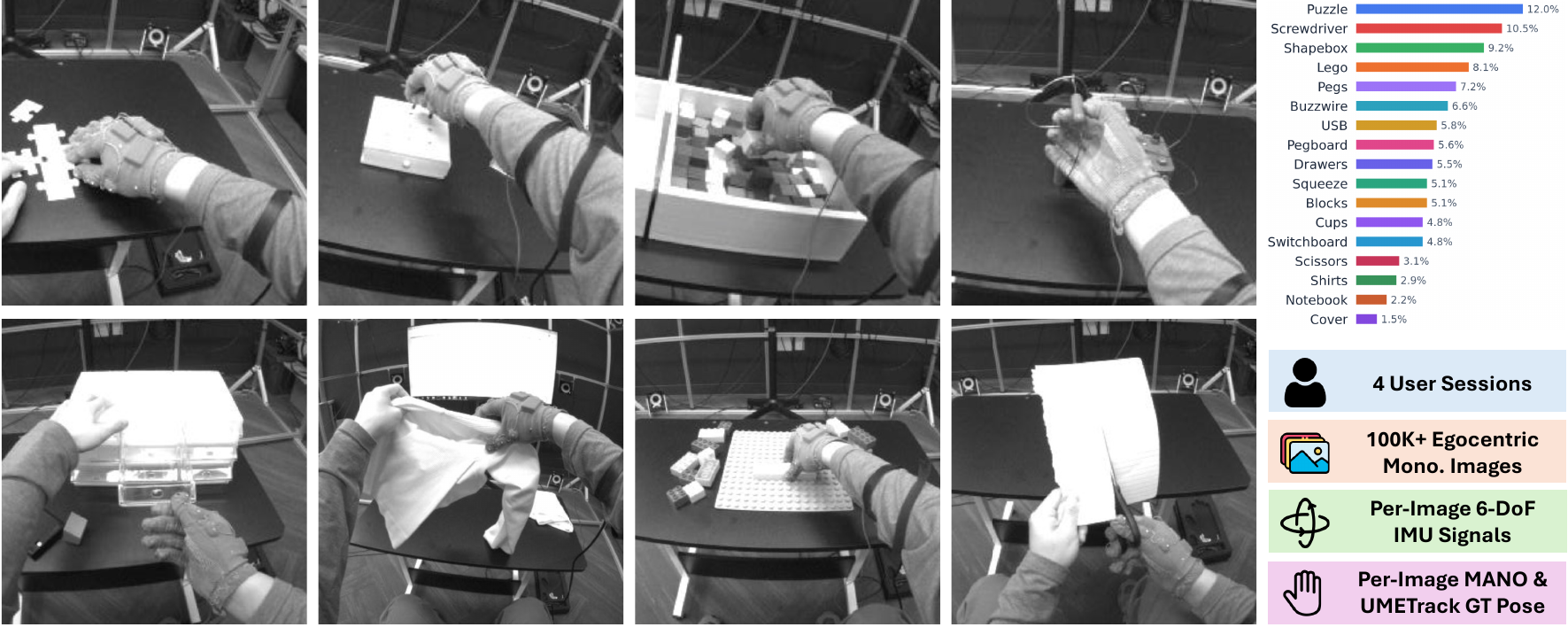}
    \caption{\textbf{Overview of DexGloveHOI dataset.} \emph{(Left)} Sample egocentric images showing diverse activities captured with the data sensing glove. \emph{(Top right)} Distribution of activity categories in DexGloveHOI. \emph{(Bottom right)} Key statistics of dataset. Each frame is paired with synchronized on-glove IMU signals and ground truth from MoCap system in both MANO and UMETrack representations.}
    \label{fig:data}
\end{figure}

\textbf{Hierarchical cross-sensor attention.} The core of AVI-HT is a cross-attention module that fuses vision and IMU
features at two hierarchical levels. We generate a global visual  
  token $\mathbf{F}_{\mathrm{vis}}^* \in \mathbb{R}^{d}$ from the visual encoder and aggregate the gravity and gyroscope features per IMU sensor, yielding a set of $N_s{=}12$ sensor tokens
   $\{\mathbf{s}_k\}_{k=1}^{N_s}$. Both token sets are projected to a shared embedding dimension $d$ and concatenated into a unified sequence $\mathbf{Z} = [\mathbf{F}_{\mathrm{vis}}^*,       
  \mathbf{s}_1, \ldots, \mathbf{s}_{N_s}] \in \mathbb{R}^{(1 + N_s) \times d}$.
Since the human hand is a kinematic chain with known topology, not all sensor--sensor interactions are equally informative: an IMU mounted on the ring finger's proximal phalanx is
  highly relevant to the distal phalanx but carries little direct information about the thumb. We encode this inductive bias via a kinematic prior mask $\mathbf{M} \in \mathbb{R}^{(1+N_s)
   \times (1+N_s)}$, where $M_{ij} = -\alpha , d_{\mathrm{geo}}(i,j)$ is the negated geodesic distance between tokens $i$ and $j$ on the hand skeleton graph, measured as the shortest-path
   hop count between two sensor mounting sites. The first-level attention is:
\begin{equation}
    \mathrm{Attention}^{\text{[1]}}(\mathbf{Q}, \mathbf{K}, \mathbf{V}) =
    \mathrm{softmax}\!\left(\frac{\mathbf{Q}\mathbf{K}^\top}{\sqrt{d_k}}
    + \mathbf{M}\right) \mathbf{V},
    \label{eq:attention}
\end{equation}
where $\mathbf{Q} = \mathbf{Z}\mathbf{W}^Q$, $\mathbf{K} = \mathbf{Z}\mathbf{W}^K$,
$\mathbf{V} = \mathbf{Z}\mathbf{W}^V$. When a hand joint is visually occluded, its attention weight
  shifts toward the spatially corresponding IMU sensor, which provides adaptive, occlusion-aware fusion that is anatomically grounded by the kinematic prior. To assess the relative importance 
  of each IMU sensor, we extract the first row of the attention matrix, $\mathbf{a}_{\mathrm{vis}} = \mathrm{softmax}(\cdot)\{1,:\} \in \mathbb{R}^{N_s}$, which captures how strongly the
  visual representation attends to each sensor under the current observation.
However, one potential limitation of the first-level attention is that the single visual token is outnumbered by $N_s{=}12$ IMU tokens, and this imbalance can cause the softmax mass to          
  concentrate on the more numerous inertial modality. To mitigate this, we introduce a second-level self-attention that operates on a balanced token pair. Specifically, we average-pool
  the $N_s$ sensor tokens from the first level into a single aggregated IMU token $\bar{\mathbf{s}} \in \mathbb{R}^{d}$ and    
  form a compact sequence $\mathbf{Z}' = [\tilde{\mathbf{F}}_{\mathrm{vis}}^*, \bar{\mathbf{s}}] \in \mathbb{R}^{2 \times d}$. A standard self-attention layer then re-calibrates the
  relative contribution of vision and inertial sensing. 
Finally, the two resulting tokens are mean-pooled into a
   global hand representation in $d$ dimension, which is passed to the respective regression heads.

\textbf{Output and Losses.} For \textbf{AVI-HT-UME}, the
regression head decodes joint angles $\phi \in \mathbb{R}^{22}$, wrist
transformation $(\mathbf{R}, \mathbf{t}) \in SE(3)$, and per-landmark
uncertainty $\sigma_\ell$. 3D landmarks are recovered via differentiable
forward kinematics. The primary loss is a landmark NLL
that jointly supervises position accuracy and uncertainty calibration supplemented by a joint angle loss \cite{han2022umetrack}.
For \textbf{AVI-HT-MANO}, the regression head decodes the fused joint tokens
into MANO pose $\theta$, shape $\beta$, and camera $\pi$ via linear
projections. Training uses a combination of 3D and 2D losses and an
adversarial loss
from a discriminator trained on hand shape, pose, and per-joint rotations
separately~\cite{pavlakos2024reconstructing}.
\section{Evaluation Dataset}
\label{sec:eval_data}

A key challenge in advancing multi-modal hand tracking is the lack of
suitable evaluation data. Existing 3D hand pose
datasets~\cite{zimmermann2019freihand,hampali2020honnotate} are
predominantly vision-only, captured in controlled studio settings \cite{moon2020interhand2}, and
limited to a narrow range of hand poses \cite{zhang2025vcot}. They do not include inertial
sensor data, nor do they reflect the diversity of hand behaviors
encountered in real-world AR/VR use, such as such as fine-grained tool manipulation. Without such data,
it is difficult to assess whether fusing vision with IMU signals
genuinely improves tracking, or merely adds complexity.

To address this gap, we collect a multi-modal evaluation dataset,
\textbf{DexGloveHOI}, that pairs synchronized egocentric video with dense
on-glove IMU signals and high-fidelity marker-based motion capture ground
truth. DexGloveHOI is designed with two goals: (1)~to cover the full
spectrum of dexterous hand use that an AR/VR or robotic system must handle, and
(2)~to provide the multi-modal signals needed to evaluate vision-IMU
fusion approaches.


We design the capture protocol to span the breadth of hand behaviors relevant to AR/VR input that are organized into 17 distinct dexterous manipulation tasks per session across several complementary categories. We show the full list of the task names and sampled egocentric images in Figure \ref{fig:data}.
Task instructions are recorded on video before each activity, and task ordering is determined per session. Each session lasts approximately 45-55 minutes of usable capture across 4 participants. Before the formal data collection, we conduct calibration for MoCap markers and IMU orientation. 
We finally collected 3.5 hours of quality-controlled
multi-modal data. This yields over 100K+
egocentric camera frames, 2.4M IMU samples, and 720\,GB of synchronized
sensor recordings.
\begin{table}[t]\centering
\caption{Ablation of fusion design for AVI-HT-UME on DexGloveHOI.}
\label{tab:fusion_ablation}
\small
\begin{tabular}{lcccccc}\toprule
\textbf{Model} & \textbf{MKPE $\downarrow$} & \textbf{F.MKPE $\downarrow$} & \textbf{MKPE.T $\downarrow$} & \textbf{F.MKPE.T $\downarrow$} & \textbf{P.A $\uparrow$} & \textbf{P.A.T $\uparrow$} \\\midrule
UMETrack                  & 12.342 & 17.478 & 9.359 & 17.798 & 76.059 & 81.433 \\\midrule
AVI-HT (early concat)     & 11.928 & 16.935 & 8.649 & 16.545 & 76.208 & 82.629 \\
AVI-HT (late fusion)      & 12.139 & 16.492 & 7.054 & 13.233 & 75.818 & 85.825 \\
AVI-HT (pooled IMU token) & 11.956 & 14.571 & 7.940 & 14.427 & 76.213 & 84.027 \\
AVI-HT (fusion-1 only)    & 11.113 & 14.088 & 7.456 & 13.740 & 77.818 & 84.994 \\
AVI-HT (w/o k-prior)      & 10.901 & 13.898 & 7.406 & 13.611 & 78.238 & 85.090 \\
AVI-HT (full)             & \textbf{10.359} & \textbf{13.253} & \textbf{7.021} & \textbf{12.785} & \textbf{79.315} & \textbf{85.859} \\
\bottomrule
\end{tabular}
\end{table}

\begin{figure}[t]
    \centering

\includegraphics[width=\linewidth]{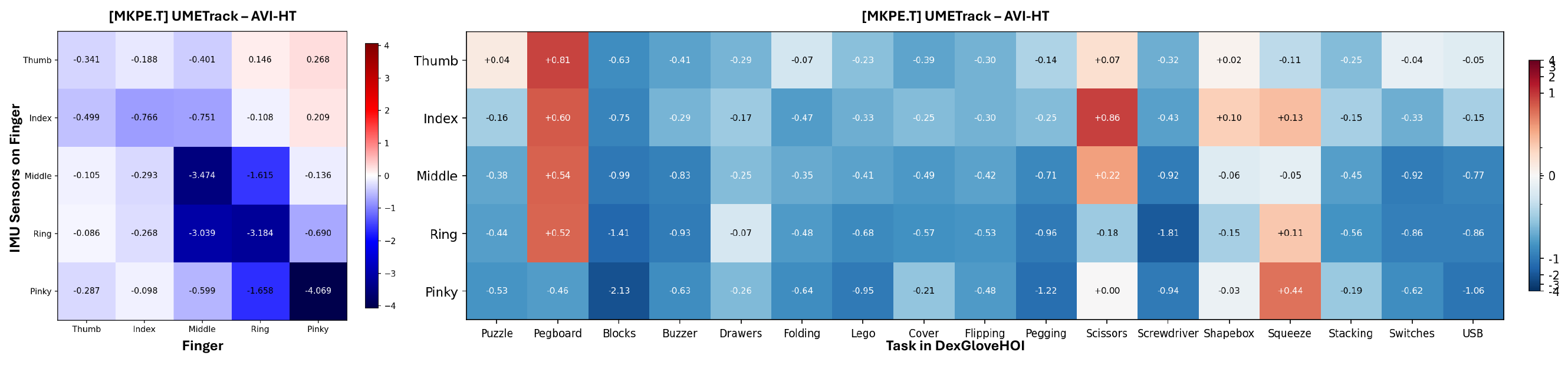}
    \caption{\textbf{Per-finger IMU contribution.} \emph{(Left)} MKPE.T Gap when only the corresponding IMU sensor group (row) is provided during training. \emph{(Right)} MKPE.T Gap broken down by activity type.}
    \label{fig:imu_ablation}
\end{figure}

\section{Experiments}
\label{sec:experiments}

We first ablate the fusion design and the contribution of the glove IMU sensors across fingers, activities, occlusion levels, and subjects (\ref{sec:exp_ablation}).
We then compare 3D hand tracking accuracy against single- and multi-modal baselines under both UMETrack and MANO hand representations (\ref{sec:exp_accuracy}). We further evaluate whether the tracked hand preserves hand-object contacts accurately and consistently over time (\ref{sec:exp_contact}). We finally study the model's sensitivity to IMU signal quality in Appendix~\ref{sec:exp_sensitivity}.



\paragraph{Implementation Details} AVI-HT is trained on a separate ${\sim}$8M-image glove dataset, and DexGloveHOI is used for evaluation only, with subjects, sessions, and tasks disjoint from training. AVI-HT-UME and AVI-HT-MANO are initialized from pretrained UMETrack and HaMeR, and these baselines are fine-tuned on the same training set with the same backbones (vision-only). Training details are in Appendix~\ref{sec:impl_details}.

\subsection{Ablation Study for IMU Sensors}
\label{sec:exp_ablation}
We ablate AVI-HT along four axes to understand which components of the multi-modal design drive its improvements: the fusion architecture, the contribution of individual IMU sensors and activity types, the severity of visual occlusion, and generalization across subjects.

\paragraph{Fusion design.}
Table~\ref{tab:fusion_ablation} compares AVI-HT-UME with simple fusion baselines and component removals: \textbf{early concat} (concatenated backbone features with a linear projection, no attention), \textbf{late fusion} (a separate IMU decoder whose joint angles are fused at the output), \textbf{pooled IMU token} (average-pooling over IMU timesteps instead of the last one), \textbf{fusion-1 only} (no second-level attention), and \textbf{w/o k-prior} (no kinematic prior mask $\mathbf{M}$). Naive fusion barely improves the global MKPE over UMETrack, since indiscriminately mixing in noisy IMU features hurts when vision alone suffices, whereas attention raises the IMU weight only when occlusion makes it beneficial. Removing the kinematic prior or the second-level attention degrades all metrics (MKPE $+0.54$\,mm and $+0.75$\,mm), showing that the gains stem from the hierarchical, topology-aware design rather than from added IMU features or parameters.

\begin{table}[t]\centering
\caption{Per-subject results of AVI-HT-UME on DexGloveHOI.}
\label{tab:per_subject}
\small
\begin{tabular}{lcccccc}\toprule
\textbf{Subject} & \textbf{MKPE $\downarrow$} & \textbf{F.MKPE $\downarrow$} & \textbf{MKPE.T $\downarrow$} & \textbf{F.MKPE.T $\downarrow$} & \textbf{P.A $\uparrow$} & \textbf{P.A.T $\uparrow$} \\\midrule
Subject 1 & 10.847 & 13.841 & 7.069 & 12.792 & 78.324 & 85.762 \\
Subject 2 &  9.624 & 12.190 & 6.548 & 12.469 & 80.797 & 86.811 \\
Subject 3 & 11.266 & 14.448 & 7.722 & 13.814 & 77.486 & 84.450 \\
Subject 4 &  8.852 & 11.512 & 6.266 & 11.311 & 82.333 & 87.371 \\
\bottomrule
\end{tabular}
\end{table}

\begin{table}[t]\centering
\caption{Fingertip error binned by the number of occluded keypoints per frame.}
\label{tab:occlusion}
\small
\begin{tabular}{lcccccc}\toprule
& \multicolumn{3}{c}{\textbf{F.MKPE $\downarrow$}} & \multicolumn{3}{c}{\textbf{F.MKPE.T $\downarrow$}} \\
\cmidrule(lr){2-4}\cmidrule(lr){5-7}
\textbf{Occluded keypoints} & UMETrack & AVI-HT & Gap & UMETrack & AVI-HT & Gap \\\midrule
$[0, 5)$   & 16.525 & 13.034 & 3.491 & 16.532 & 12.605 & 3.927 \\
$[5, 10)$  & 16.704 & 13.226 & 3.478 & 17.383 & 12.745 & 4.638 \\
$[10, 15)$ & 17.835 & 13.373 & 4.462 & 18.744 & 12.936 & 5.808 \\
$[15, 20)$ & 21.741 & 15.539 & 6.202 & 21.067 & 13.585 & 7.482 \\
\bottomrule
\end{tabular}
\end{table}

\paragraph{Per-subject generalization.}
Table~\ref{tab:per_subject} reports AVI-HT-UME results for each of the four held-out DexGloveHOI subjects. AVI-HT performs consistently across subjects with small variance, and every subject outperforms the aggregate vision-only UMETrack result (Table~\ref{global_ume_acc}) on all six metrics. Since no DexGloveHOI subject is seen during training, this indicates that the gains come from vision-IMU fusion rather than subject-specific overfitting.

\paragraph{Per-finger and per-activity contribution.}
Figure~\ref{fig:imu_ablation} reports the \emph{MKPE.T Gap} (AVI-HT $-$ UMETrack; negative is better). The left heatmap shows \textbf{per-finger IMU contribution}: when only one IMU sensor group (row) is provided, the largest improvement consistently falls on its own finger (diagonal). Near-diagonal cells also improve notably due to kinematic coupling, as neighboring fingers share tendons and move together. It matches our cross-sensor attention design, which routes each finger information mainly through its sensor while still exploiting cross-finger cues.

The right heatmap shows \textbf{per-activity-type breakdown}: for each
IMU sensor group (rows) and activity category (columns), we report
MKPE.T improvement with the same AVI-HT $-$ UMETrack metric. Activities involving heavy occlusion and fine-grained manipulation (e.g., cutting, scissors, screwdriver) show
the largest improvements, which confirms that IMU fusion provides the
greatest benefit precisely in the scenarios where vision alone struggles
most.

\paragraph{Occlusion severity.}
To directly quantify the benefit under occlusion, we label each keypoint in each frame as occluded or visible using the camera parameters and the 3D ground-truth pose, and bin the frames by the number of occluded keypoints. Table~\ref{tab:occlusion} reports F.MKPE and F.MKPE.T, the metrics most sensitive to the finger-mounted IMUs. As occlusion increases, the UMETrack error grows by 5.2\,mm in F.MKPE, whereas AVI-HT grows by only 2.5\,mm. Consequently, the AVI-HT advantage grows monotonically with occlusion severity, from a 3.5--3.9\,mm gap in the least-occluded bin to 6.2--7.5\,mm in the most-occluded bin. This confirms that the gains are concentrated where visual evidence is lost and scale with the severity of occlusion.

\begin{table}[t]\centering
\caption{3D Hand Tracking Accuracy for UMETrack Pose}
\label{global_ume_acc}
\small
\begin{tabular}{lcccccc}\toprule
\textbf{Model} & \textbf{MKPE $\downarrow$} & \textbf{F.MKPE $\downarrow$} & \textbf{MKPE.T $\downarrow$} & \textbf{F.MKPE.T $\downarrow$} & \textbf{P.A $\uparrow$} & \textbf{P.A.T $\uparrow$} \\\midrule
UMETrack       & 12.342 & 17.478 & 9.359  & 17.798   & 76.059 & 81.433 \\
UMETrack + EKF & 15.406 & 18.658 & 7.891  & 14.232   & 70.903 & 84.210 \\
IMU-Tracker    & -      & -      & 7.151  & 12.923   & -      & 85.769 \\
AVI-HT           & \textbf{10.359} & \textbf{13.253} & \textbf{7.021}  & \textbf{12.785}   & \textbf{79.315} & \textbf{85.859} \\
\bottomrule
\end{tabular}
\end{table}

\begin{table}[t]\centering
\caption{3D Hand Tracking Accuracy for MANO Pose}
\label{mano_acc}
\small
\begin{tabular}{lcccc}\toprule
\textbf{} & \textbf{PA-MPJPE $\downarrow$} & \textbf{PA-MPVPE $\downarrow$} & \textbf{F@5 $\uparrow$} & \textbf{F@15 $\uparrow$} \\\midrule
HaMeR & 13.754   & 12.736   & 0.516 & 0.882 \\
AVI-HT  & \textbf{10.519}   & \textbf{9.265}    & \textbf{0.628} & \textbf{0.936 }\\
\bottomrule
\end{tabular}
\end{table}

\begin{figure}[t]\centering
    \centering

\includegraphics[width=\linewidth]{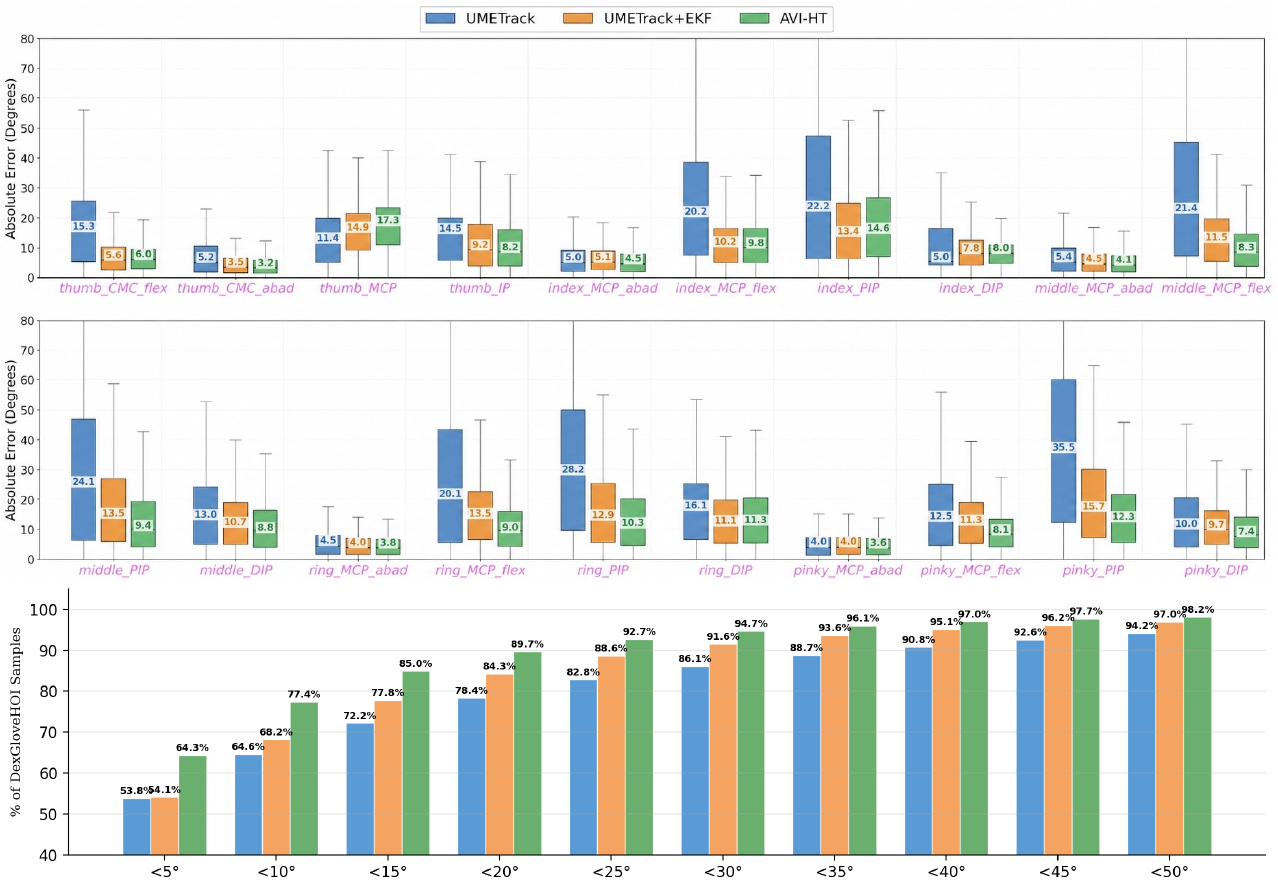}
    \caption{\textbf{Breakdown evaluation for AVI-HT.} \emph{(Top)} Absolute angle error across all 22 degrees of freedom for \textcolor{myblue}{UMETrack}, \textcolor{myorange}{UMETrack + EKF}, and \textcolor{mygreen}{AVI-HT}. AVI-HT reduces error on most joints, with the largest gains on MCP flexion joints. \emph{(Bottom)} Cumulative distribution of joint angle errors over all DexGloveHOI samples. AVI-HT achieves a higher fraction of samples below each error threshold.}
    \label{fig:joint_breakdown}
\end{figure}

\subsection{3D Hand Tracking Accuracy}
\label{sec:exp_accuracy}

\paragraph{UMETrack hand model.}
We compare AVI-HT on DexGloveHOI with three baselines that are finetuned on the same vision dataset:
\textbf{UMETrack}~\cite{han2022umetrack} (vision-only),
\textbf{UMETrack + EKF} (vision-IMU fusion by a post-hoc Extended Kalman Filter)~\cite{lei2023novel}, and \textbf{IMU-Tracker}
(6-DoF IMU-only)~\cite{sarker2026real}. Table~\ref{global_ume_acc} reports the mean
keypoint position error (MKPE), fingertip-only MKPE (F.MKPE), their variants aligned at the GT hand root (\textbf{T}ransformed; MKPE.T, F.MKPE.T), and PCK AUC (P.A, P.A.T).
AVI-HT outperforms all baselines on every metric. It achieves the lowest MKPE and F.MKPE, improving over
UMETrack by 16.1\% and 24.2\%. AVI-HT also
outperforms UMETrack + EKF, showing that learned cross-sensor
attention fuses the signals more effectively than a hand-crafted post-hoc
filter. IMU-Tracker achieves competitive fingertip accuracy but cannot estimate the global wrist
position from 6-DoF IMUs, which highlights the complementary vision-IMU.

Figure~\ref{fig:joint_breakdown} (top) breaks down the absolute joint angle error over all 22 degrees
of freedom. AVI-HT reduces the error relative
to both UMETrack and UMETrack + EKF, with the largest
gains on the MCP flexion joints that are most prone to occlusion. The cumulative error distribution (bottom) shows that AVI-HT places a larger fraction of samples below every threshold, with the largest gap at strict thresholds such as ${<}5^\circ$, which matters for high-precision dexterous manipulation.

\paragraph{MANO hand model.}
We compare AVI-HT with the finetuned
HaMeR~\cite{pavlakos2024reconstructing}, a state-of-the-art
vision-only hand mesh recovery method. Table~\ref{mano_acc}
reports Procrustes-aligned mean per-joint and per-vertex position errors (PA-MPJPE, PA-MPVPE) and
F-scores at 5\,mm and 15\,mm. AVI-HT reduces PA-MPJPE
from 13.754\,mm to 10.519\,mm (23.5\%) and PA-MPVPE from
12.736\,mm to 9.265\,mm (27.3\%), and raises F@5 from
0.516 to 0.628 and F@15 from 0.882 to 0.936, showing that the fusion substantially improves even a strong
transformer-based baseline.

\begin{figure}[t]
    \centering

\includegraphics[width=0.9\linewidth]{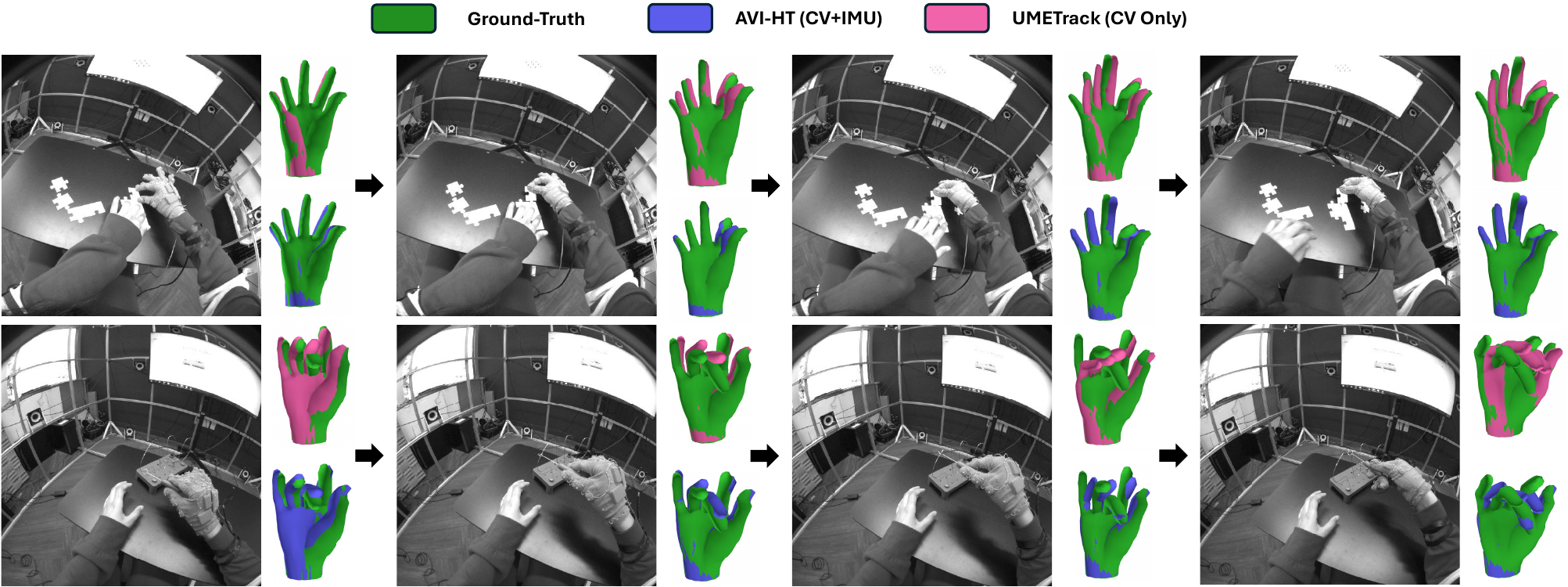}
    \caption{\textbf{Qualitative tracking comparison.} Two hand-object interaction sequences with finger occlusion moments. AVI-HT closely tracks the ground truth under heavy occlusion, while UMETrack deviates at unseen fingers from egocentric views.}
    \label{fig:vis}
\end{figure}

\paragraph{Qualitative results.}
Figure~\ref{fig:vis} compares the MoCap ground truth, AVI-HT and UMETrack on two
egocentric hand-object interaction sequences in which finger occlusion
varies during manipulation. Both methods perform similarly when all fingers are visible. As fingers become occluded, AVI-HT still tracks the ground truth closely, whereas UMETrack deviates
substantially on the unseen fingers, showing that the cross-sensor attention leverages IMU cues when visual evidence is lost. More examples and tracking videos with attention scores are in Appendix~\ref{more_vis}.

\begin{table}[t]\centering
\caption{3D Hand Tracking Accuracy for MANO Pose}
\label{mano_acc}
\small
\begin{tabular}{lcccc}\toprule
\textbf{} & \textbf{PA-MPJPE $\downarrow$} & \textbf{PA-MPVPE $\downarrow$} & \textbf{F@5 $\uparrow$} & \textbf{F@15 $\uparrow$} \\\midrule
HaMeR & 13.754   & 12.736   & 0.516 & 0.882 \\
AVI-HT  & \textbf{10.519}   & \textbf{9.265}    & \textbf{0.628} & \textbf{0.936 }\\
\bottomrule
\end{tabular}
\end{table}

\subsection{Hand-Object Contact Evaluation}
\label{sec:exp_contact}
\begin{figure}[h!]
    \centering
\includegraphics[width=\linewidth]{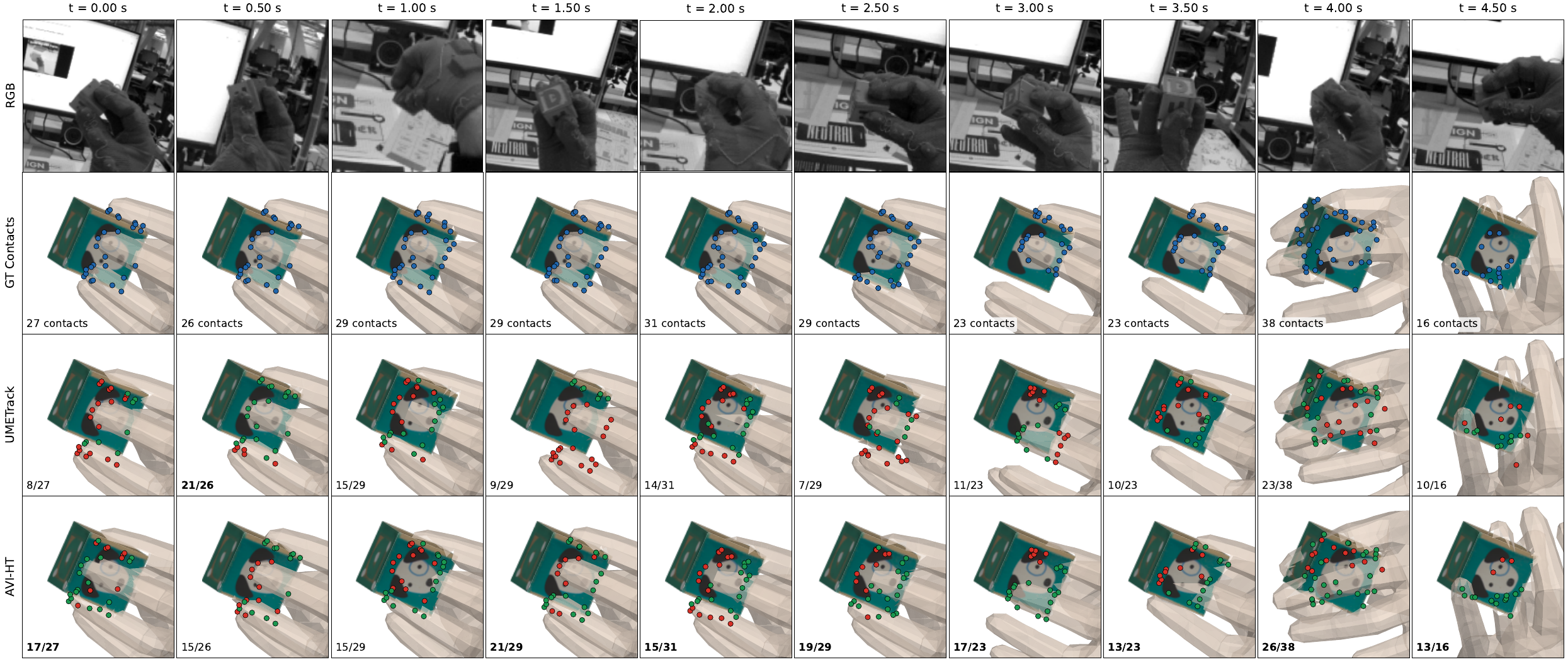}
    \caption{\textbf{Hand-object contact over time} (0.5\,s apart). Ground-truth contacts (blue) are green in each method's row if kept within 5\,mm and red if missed; X/Y counts kept contacts.}
    \label{fig:contact_vis}
\end{figure}

Keypoint errors do not show whether the fingers stay on a grasped object. We therefore evaluate hand-object contact on 15 wooden-block sessions from the DexGloveHOI dataset, with the block pose from MoCap. With the predicted wrist aligned to the ground truth, a hand vertex within 2\,mm of the block is a ground-truth contact, and it is kept if the predicted vertex lies within $\tau$ of the block. \textbf{Spatial} metrics are per frame: contact displacement (Disp.) on the block surface and retention (Ret@$\tau$), the fraction of contacts kept. \textbf{Temporal} metrics follow each contact lasting at least 0.5\,s: the fraction of time kept (Capt.), the longest uninterrupted fraction (Longest), and losses per contact-vertex second.
As shown in Table~\ref{tab:contact}, AVI-HT improves every metric, reducing displacement by 0.95\,mm and contact losses by 8.5\% (29.7\% at $\tau = 10$\,mm). Figure~\ref{fig:contact_vis} shows that UMETrack repeatedly loses occluded fingertip contacts that AVI-HT keeps.

\begin{table}[t]\centering
\caption{Hand-object contact evaluation. Temporal metrics use $\tau = 5$\,mm. Disp.\ is in mm and Loss/s in losses per second; others in \%.}
\label{tab:contact}
\small
\begin{tabular}{lcccccc}\toprule
& \multicolumn{3}{c}{\textbf{Spatial}} & \multicolumn{3}{c}{\textbf{Temporal}} \\
\cmidrule(lr){2-4}\cmidrule(lr){5-7}
\textbf{Model} & \textbf{Disp. $\downarrow$} & \textbf{Ret@5 $\uparrow$} & \textbf{Ret@10 $\uparrow$} & \textbf{Capt. $\uparrow$} & \textbf{Longest $\uparrow$} & \textbf{Loss/s $\downarrow$} \\\midrule
UMETrack & 9.91 & 56.60 & 84.12 & 58.09 & 50.13 & 0.911 \\
AVI-HT & \textbf{8.96} & \textbf{59.41} & \textbf{86.73} & \textbf{61.20} & \textbf{53.84} & \textbf{0.834} \\
\bottomrule
\end{tabular}
\end{table}

\section{Conclusion}
We presented AVI-HT, an adaptive vision-IMU fusion approach for 3D hand
tracking that introduces cross-sensor attention to fuse
egocentric visual features with on-glove IMU signals. The design is general across vision backbones
and hand models.
Experiment results on DexGloveHOI show consistent
improvements over various baselines on both UMETrack and MANO hand models, with
the largest gains under heavy occlusion. Ablation study reveals that each IMU sensor primarily
benefits its own finger with additional cross-finger transfer from kinematic coupling, and sensitivity studies confirm certain robustness to IMU
noise and temporal misalignment.

\paragraph{Limitations.}
The data sensing glove changes the hand's appearance, which may cause a domain gap for downstream pipelines that expect bare-hand images. Hand avatar methods \cite{chen2023handavatar, zhang2026glove2hand} could mitigate this by rendering a photorealistic bare hand from the tracked pose. In addition, we use one type of glove, so generalization to other sensor layouts, form factors, and IMU specifications remains to be validated, motivating future work on glove-agnostic fusion.





\bibliographystyle{plain}
\bibliography{references}

@inproceedings{girdhar2023imagebind,
  title={Imagebind: One embedding space to bind them all},
  author={Girdhar, Rohit and El-Nouby, Alaaeldin and Liu, Zhuang and Singh, Mannat and Alwala, Kalyan Vasudev and Joulin, Armand and Misra, Ishan},
  booktitle={Proceedings of the IEEE/CVF conference on computer vision and pattern recognition},
  pages={15180--15190},
  year={2023}
}

@inproceedings{bachmann2022multimae,
  title={Multimae: Multi-modal multi-task masked autoencoders},
  author={Bachmann, Roman and Mizrahi, David and Atanov, Andrei and Zamir, Amir},
  booktitle={European conference on computer vision},
  pages={348--367},
  year={2022},
  organization={Springer}
}

@inproceedings{pan2023fusing,
  title={Fusing monocular images and sparse imu signals for real-time human motion capture},
  author={Pan, Shaohua and Ma, Qi and Yi, Xinyu and Hu, Weifeng and Wang, Xiong and Zhou, Xingkang and Li, Jijunnan and Xu, Feng},
  booktitle={SIGGRAPH Asia 2023 Conference Papers},
  pages={1--11},
  year={2023}
}

@article{bao2022fusepose,
  title={FusePose: IMU-vision sensor fusion in kinematic space for parametric human pose estimation},
  author={Bao, Yiming and Zhao, Xu and Qian, Dahong},
  journal={IEEE Transactions on Multimedia},
  volume={25},
  pages={7736--7746},
  year={2022},
  publisher={IEEE}
}

@inproceedings{pavlakos2024reconstructing,
  title={Reconstructing hands in 3d with transformers},
  author={Pavlakos, Georgios and Shan, Dandan and Radosavovic, Ilija and Kanazawa, Angjoo and Fouhey, David and Malik, Jitendra},
  booktitle={Proceedings of the IEEE/CVF Conference on Computer Vision and Pattern Recognition},
  pages={9826--9836},
  year={2024}
}

@inproceedings{rong2021frankmocap,
  title={Frankmocap: A monocular 3d whole-body pose estimation system via regression and integration},
  author={Rong, Yu and Shiratori, Takaaki and Joo, Hanbyul},
  booktitle={Proceedings of the IEEE/CVF International Conference on Computer Vision},
  pages={1749--1759},
  year={2021}
}

@inproceedings{banerjee2025hot3d,
  title={Hot3d: Hand and object tracking in 3d from egocentric multi-view videos},
  author={Banerjee, Prithviraj and Shkodrani, Sindi and Moulon, Pierre and Hampali, Shreyas and Han, Shangchen and Zhang, Fan and Zhang, Linguang and Fountain, Jade and Miller, Edward and Basol, Selen and others},
  booktitle={Proceedings of the IEEE/CVF Conference on Computer Vision and Pattern Recognition},
  pages={7061--7071},
  year={2025}
}

@inproceedings{zimmermann2017learning,
  title={Learning to estimate 3d hand pose from single rgb images},
  author={Zimmermann, Christian and Brox, Thomas},
  booktitle={Proceedings of the IEEE international conference on computer vision},
  pages={4903--4911},
  year={2017}
}

@inproceedings{arunachalam2023dexterous,
  title={Dexterous imitation made easy: A learning-based framework for efficient dexterous manipulation},
  author={Arunachalam, Sridhar Pandian and Silwal, Sneha and Evans, Ben and Pinto, Lerrel},
  booktitle={2023 ieee international conference on robotics and automation (icra)},
  pages={5954--5961},
  year={2023},
  organization={IEEE}
}

@article{qin2023anyteleop,
  title={Anyteleop: A general vision-based dexterous robot arm-hand teleoperation system},
  author={Qin, Yuzhe and Yang, Wei and Huang, Binghao and Van Wyk, Karl and Su, Hao and Wang, Xiaolong and Chao, Yu-Wei and Fox, Dieter},
  journal={arXiv preprint arXiv:2307.04577},
  year={2023}
}

@article{zhang2020mediapipe,
  title={Mediapipe hands: On-device real-time hand tracking},
  author={Zhang, Fan and Bazarevsky, Valentin and Vakunov, Andrey and Tkachenka, Andrei and Sung, George and Chang, Chuo-Ling and Grundmann, Matthias},
  journal={arXiv preprint arXiv:2006.10214},
  year={2020}
}

@inproceedings{moon2020interhand2,
  title={Interhand2. 6m: A dataset and baseline for 3d interacting hand pose estimation from a single rgb image},
  author={Moon, Gyeongsik and Yu, Shoou-I and Wen, He and Shiratori, Takaaki and Lee, Kyoung Mu},
  booktitle={European Conference on Computer Vision},
  pages={548--564},
  year={2020},
  organization={Springer}
}

@article{lei2023novel,
  title={A novel sensor fusion approach for precise hand tracking in virtual reality-based human—Computer interaction},
  author={Lei, Yu and Deng, Yi and Dong, Lin and Li, Xiaohui and Li, Xiangnan and Su, Zhi},
  journal={Biomimetics},
  volume={8},
  number={3},
  pages={326},
  year={2023},
  publisher={MDPI}
}

@article{zhang2025vcot,
  title={VCoT-Grasp: Grasp Foundation Models with Visual Chain-of-Thought Reasoning for Language-driven Grasp Generation},
  author={Zhang, Haoran and Bai, Shuanghao and Zhou, Wanqi and Zhang, Yuedi and Zhang, Qi and Ding, Pengxiang and Chi, Cheng and Wang, Donglin and Chen, Badong},
  journal={arXiv preprint arXiv:2510.05827},
  year={2025}
}

@incollection{lewis2023pose,
  title={Pose space deformation: a unified approach to shape interpolation and skeleton-driven deformation},
  author={Lewis, John P and Cordner, Matt and Fong, Nickson},
  booktitle={Seminal Graphics Papers: Pushing the Boundaries, Volume 2},
  pages={811--818},
  year={2023}
}

@article{laidig2023vqf,
  title={VQF: Highly accurate IMU orientation estimation with bias estimation and magnetic disturbance rejection},
  author={Laidig, Daniel and Seel, Thomas},
  journal={Information Fusion},
  volume={91},
  pages={187--204},
  year={2023},
  publisher={Elsevier}
}

@inproceedings{boukhayma20193d,
  title={3d hand shape and pose from images in the wild},
  author={Boukhayma, Adnane and Bem, Rodrigo de and Torr, Philip HS},
  booktitle={Proceedings of the IEEE/CVF conference on computer vision and pattern recognition},
  pages={10843--10852},
  year={2019}
}

@inproceedings{chen2023handavatar,
  title={Hand Avatar: Free-Pose Hand Animation and Rendering from Monocular Video},
  author={Chen, Xingyu and Wang, Baoyuan and Shum, Heung-Yeung},
  booktitle={Proceedings of the IEEE/CVF Conference on Computer Vision and Pattern Recognition},
  pages={8683--8693},
  year={2023}
}

@inproceedings{lourakis2005levenberg,
  title={Is Levenberg-Marquardt the most efficient optimization algorithm for implementing bundle adjustment?},
  author={Lourakis, Manolis LA and Argyros, Antonis A},
  booktitle={Tenth IEEE International Conference on Computer Vision (ICCV'05) Volume 1},
  volume={2},
  pages={1526--1531},
  year={2005},
  organization={IEEE}
}

@article{MANO:SIGGRAPHASIA:2017,
      title = {Embodied Hands: Modeling and Capturing Hands and Bodies Together},
      author = {Romero, Javier and Tzionas, Dimitrios and Black, Michael J.},
      journal = {ACM Transactions on Graphics, (Proc. SIGGRAPH Asia)},
      volume = {36},
      number = {6},
      series = {245:1--245:17},
      month = nov,
      year = {2017},
      month_numeric = {11}
  }

@inproceedings{sarker2026real,
  title={Real-Time Hand Pose Tracking using 6-Axis IMUs},
  author={Sarker, Anik and Kou, Ziyi and Ristani, Ergys and Guan, Li and Niehues, Taylor},
  booktitle={Proceedings of the 21st ACM/IEEE International Conference on Human-Robot Interaction},
  pages={1182--1191},
  year={2026}
}

@inproceedings{mueller2017real,
  title={Real-time hand tracking under occlusion from an egocentric rgb-d sensor},
  author={Mueller, Franziska and Mehta, Dushyant and Sotnychenko, Oleksandr and Sridhar, Srinath and Casas, Dan and Theobalt, Christian},
  booktitle={Proceedings of the IEEE international conference on computer vision},
  pages={1154--1163},
  year={2017}
}

@article{maereg2017low,
  title={A low-cost, wearable Opto-Inertial 6-DOF hand pose tracking system for VR},
  author={Maereg, Andualem T and Secco, Emanuele L and Agidew, Tayachew F and Reid, David and Nagar, Atulya K},
  journal={Technologies},
  volume={5},
  number={3},
  pages={49},
  year={2017},
  publisher={MDPI}
}

@inproceedings{lin2021two,
  title={Two-hand global 3d pose estimation using monocular rgb},
  author={Lin, Fanqing and Wilhelm, Connor and Martinez, Tony},
  booktitle={Proceedings of the IEEE/CVF winter conference on applications of computer vision},
  pages={2373--2381},
  year={2021}
}

@article{guo20223d,
  title={3D hand pose estimation from monocular RGB with feature interaction module},
  author={Guo, Shaoxiang and Rigall, Eric and Ju, Yakun and Dong, Junyu},
  journal={IEEE Transactions on Circuits and Systems for Video Technology},
  volume={32},
  number={8},
  pages={5293--5306},
  year={2022},
  publisher={IEEE}
}

@inproceedings{lin2021end,
  title={End-to-end human pose and mesh reconstruction with transformers},
  author={Lin, Kevin and Wang, Lijuan and Liu, Zicheng},
  booktitle={Proceedings of the IEEE/CVF conference on computer vision and pattern recognition},
  pages={1954--1963},
  year={2021}
}

@inproceedings{lin2021mesh,
  title={Mesh graphormer},
  author={Lin, Kevin and Wang, Lijuan and Liu, Zicheng},
  booktitle={Proceedings of the IEEE/CVF international conference on computer vision},
  pages={12939--12948},
  year={2021}
}

@article{li2025handnet,
  title={HandNet: Occlusion-robust 3D hand mesh reconstruction with prior information},
  author={Li, Jiawen and Jiang, Fei and Zhu, Dandan and Zhou, Aimin},
  journal={Available at SSRN 5244153},
  year={2025}
}

@inproceedings{mummadi2018real,
  title={Real-time and embedded detection of hand gestures with an IMU-based glove},
  author={Mummadi, Chaithanya Kumar and Philips Peter Leo, Frederic and Deep Verma, Keshav and Kasireddy, Shivaji and Scholl, Philipp M and Kempfle, Jochen and Van Laerhoven, Kristof},
  booktitle={Informatics},
  volume={5},
  number={2},
  pages={28},
  year={2018},
  organization={MDPI}
}

@article{huang2018deep,
  title={Deep inertial poser: Learning to reconstruct human pose from sparse inertial measurements in real time},
  author={Huang, Yinghao and Kaufmann, Manuel and Aksan, Emre and Black, Michael J and Hilliges, Otmar and Pons-Moll, Gerard},
  journal={ACM Transactions on Graphics (TOG)},
  volume={37},
  number={6},
  pages={1--15},
  year={2018},
  publisher={ACM New York, NY, USA}
}

@article{yi2021transpose,
  title={Transpose: Real-time 3d human translation and pose estimation with six inertial sensors},
  author={Yi, Xinyu and Zhou, Yuxiao and Xu, Feng},
  journal={ACM Transactions On Graphics (TOG)},
  volume={40},
  number={4},
  pages={1--13},
  year={2021},
  publisher={ACM New York, NY, USA}
}

@inproceedings{zhang2020fusing,
  title={Fusing wearable imus with multi-view images for human pose estimation: A geometric approach},
  author={Zhang, Zhe and Wang, Chunyu and Qin, Wenhu and Zeng, Wenjun},
  booktitle={Proceedings of the IEEE/CVF Conference on Computer Vision and Pattern Recognition},
  pages={2200--2209},
  year={2020}
}

@inproceedings{li2023ego,
  title={Ego-body pose estimation via ego-head pose estimation},
  author={Li, Jiaman and Liu, Karen and Wu, Jiajun},
  booktitle={Proceedings of the IEEE/CVF Conference on Computer Vision and Pattern Recognition},
  pages={17142--17151},
  year={2023}
}

@article{zhang2026glove2hand,
  title={Glove2Hand: Synthesizing Natural Hand-Object Interaction from Multi-Modal Sensing Gloves},
  author={Zhang, Xinyu and Kou, Ziyi and Qin, Chuan and Huang, Mia and Ristani, Ergys and Kumar, Ankit and Chen, Lele and He, Kun and Boularias, Abdeslam and Guan, Li},
  journal={arXiv preprint arXiv:2603.20850},
  year={2026}
}

@inproceedings{han2022umetrack,
  title={UmeTrack: Unified multi-view end-to-end hand tracking for VR},
  author={Han, Shangchen and Wu, Po-chen and Zhang, Yubo and Liu, Beibei and Zhang, Linguang and Wang, Zheng and Si, Weiguang and Zhang, Peizhao and Cai, Yujun and Hodan, Tomas and others},
  booktitle={SIGGRAPH Asia 2022 conference papers},
  pages={1--9},
  year={2022}
}

@article{han2020megatrack,
  title={Megatrack: monochrome egocentric articulated hand-tracking for virtual reality},
  author={Han, Shangchen and Liu, Beibei and Cabezas, Randi and Twigg, Christopher D and Zhang, Peizhao and Petkau, Jeff and Yu, Tsz-Ho and Tai, Chun-Jung and Akbay, Muzaffer and Wang, Zheng and others},
  journal={ACM Transactions on Graphics (ToG)},
  volume={39},
  number={4},
  pages={87--1},
  year={2020},
  publisher={ACM New York, NY, USA}
}

@article{ohkawa2023efficient,
  title={Efficient annotation and learning for 3d hand pose estimation: A survey},
  author={Ohkawa, Takehiko and Furuta, Ryosuke and Sato, Yoichi},
  journal={International Journal of Computer Vision},
  volume={131},
  number={12},
  pages={3193--3206},
  year={2023},
  publisher={Springer}
}

@article{vaswani2017attention,
  title={Attention is all you need},
  author={Vaswani, Ashish and Shazeer, Noam and Parmar, Niki and Uszkoreit, Jakob and Jones, Llion and Gomez, Aidan N and Kaiser, {\L}ukasz and Polosukhin, Illia},
  journal={Advances in neural information processing systems},
  volume={30},
  year={2017}
}

@article{dosovitskiy2020image,
  title={An image is worth 16x16 words: Transformers for image recognition at scale},
  author={Dosovitskiy, Alexey and Beyer, Lucas and Kolesnikov, Alexander and Weissenborn, Dirk and Zhai, Xiaohua and Unterthiner, Thomas and Dehghani, Mostafa and Minderer, Matthias and Heigold, Georg and Gelly, Sylvain and others},
  journal={arXiv preprint arXiv:2010.11929},
  year={2020}
}

@article{xu2022vitpose,
  title={Vitpose: Simple vision transformer baselines for human pose estimation},
  author={Xu, Yufei and Zhang, Jing and Zhang, Qiming and Tao, Dacheng},
  journal={Advances in neural information processing systems},
  volume={35},
  pages={38571--38584},
  year={2022}
}

@inproceedings{he2016deep,
  title={Deep residual learning for image recognition},
  author={He, Kaiming and Zhang, Xiangyu and Ren, Shaoqing and Sun, Jian},
  booktitle={Proceedings of the IEEE conference on computer vision and pattern recognition},
  pages={770--778},
  year={2016}
}

@inproceedings{zimmermann2019freihand,
  title={Freihand: A dataset for markerless capture of hand pose and shape from single rgb images},
  author={Zimmermann, Christian and Ceylan, Duygu and Yang, Jimei and Russell, Bryan and Argus, Max and Brox, Thomas},
  booktitle={Proceedings of the IEEE/CVF international conference on computer vision},
  pages={813--822},
  year={2019}
}

@inproceedings{hampali2020honnotate,
  title={Honnotate: A method for 3d annotation of hand and object poses},
  author={Hampali, Shreyas and Rad, Mahdi and Oberweger, Markus and Lepetit, Vincent},
  booktitle={Proceedings of the IEEE/CVF conference on computer vision and pattern recognition},
  pages={3196--3206},
  year={2020}
}

\newpage
\appendix

\section{Implementation Details}
\label{sec:impl_details}
We train AVI-HT for 40 epochs with a batch size of 48 on $8\times$ NVIDIA H100 GPUs. We use the Adam optimizer with a learning rate of $7.89\times10^{-4}$, weight decay of $10^{-6}$, momentum of $0.9$, and $\epsilon=10^{-8}$, with a step-decay scheduler that reduces the learning rate by $10\times$ at epoch 30. Data augmentation includes fast-motion augmentation, per-frame image noise injection ($\sigma=10$), and gamma jittering from $0.5$ to $1.7$. Training takes around 24.5 hours for AVI-HT-UME and 42 hours for AVI-HT-MANO.

\section{Marker Based Mocap System Setup}
\label{mocap}
\begin{figure}[h]
    \centering

\includegraphics[width=0.5\linewidth]{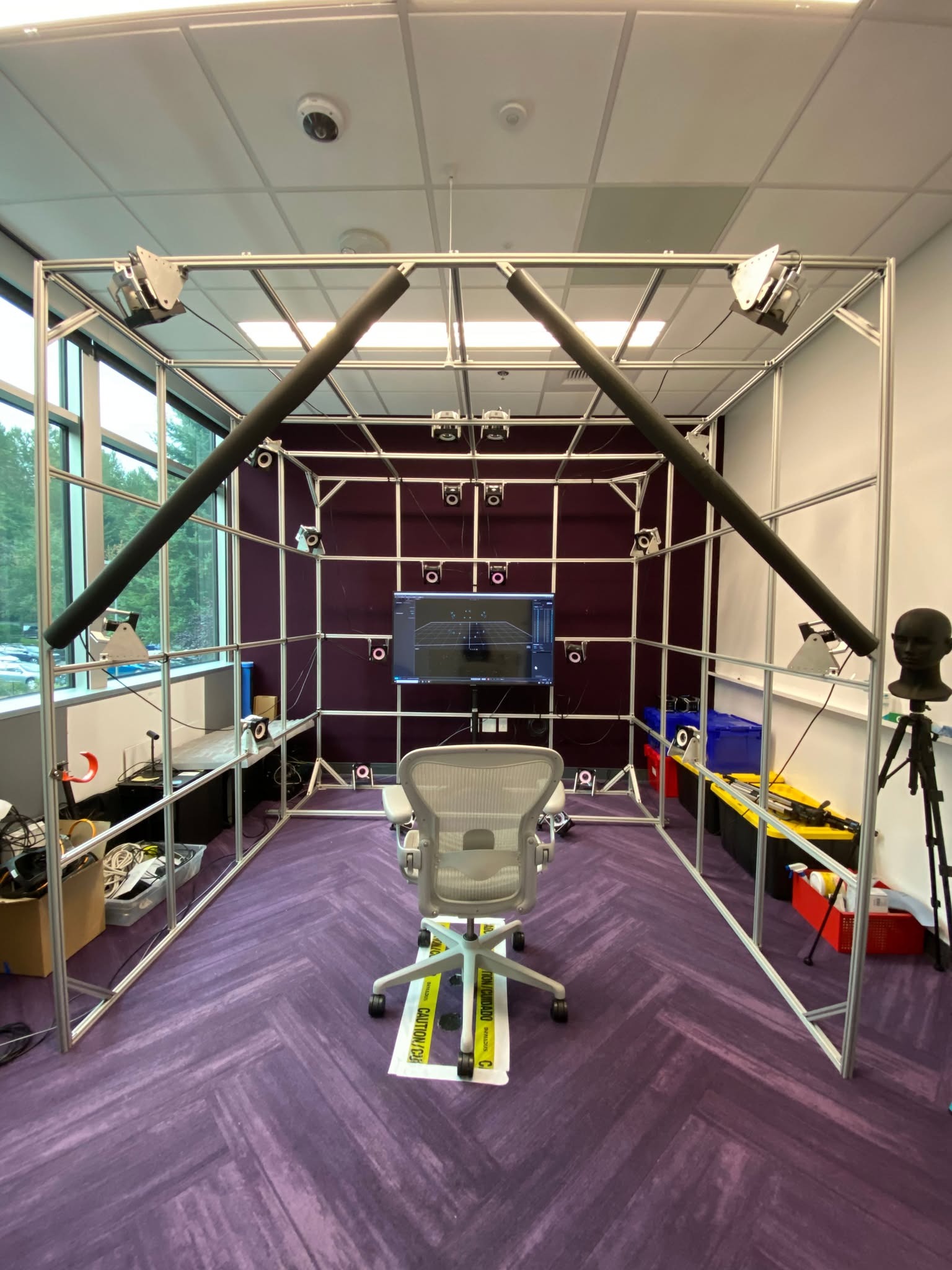}
    \caption{Marker Based Mocap System Setup}
    \label{fig:mocap_setup}
\end{figure}

To obtain ground-truth 3D hand pose annotations, we construct a dedicated marker-based optical motion capture environment. As shown in Figure \ref{fig:mocap_setup}, the capture volume is defined by a modular aluminum frame structure. We mount 12 infrared cameras at varying heights and angles around the frame to ensure dense, multi-view coverage of the capture volume and minimize marker occlusion. The back wall is covered with a dark matte backdrop to reduce infrared reflections that could introduce spurious marker detections. Subjects are seated in a fixed chair at the center of the capture volume, with the designated capture area demarcated on the floor. This seated configuration constrains the hand workspace to a consistent region within the calibrated volume, maximizing marker visibility across all cameras and ensuring reliable sub-millimeter tracking accuracy throughout data collection.

\section{More Visualization for AVI-HT}
\label{more_vis}
\begin{figure}[h]
    \centering

\includegraphics[width=\linewidth]{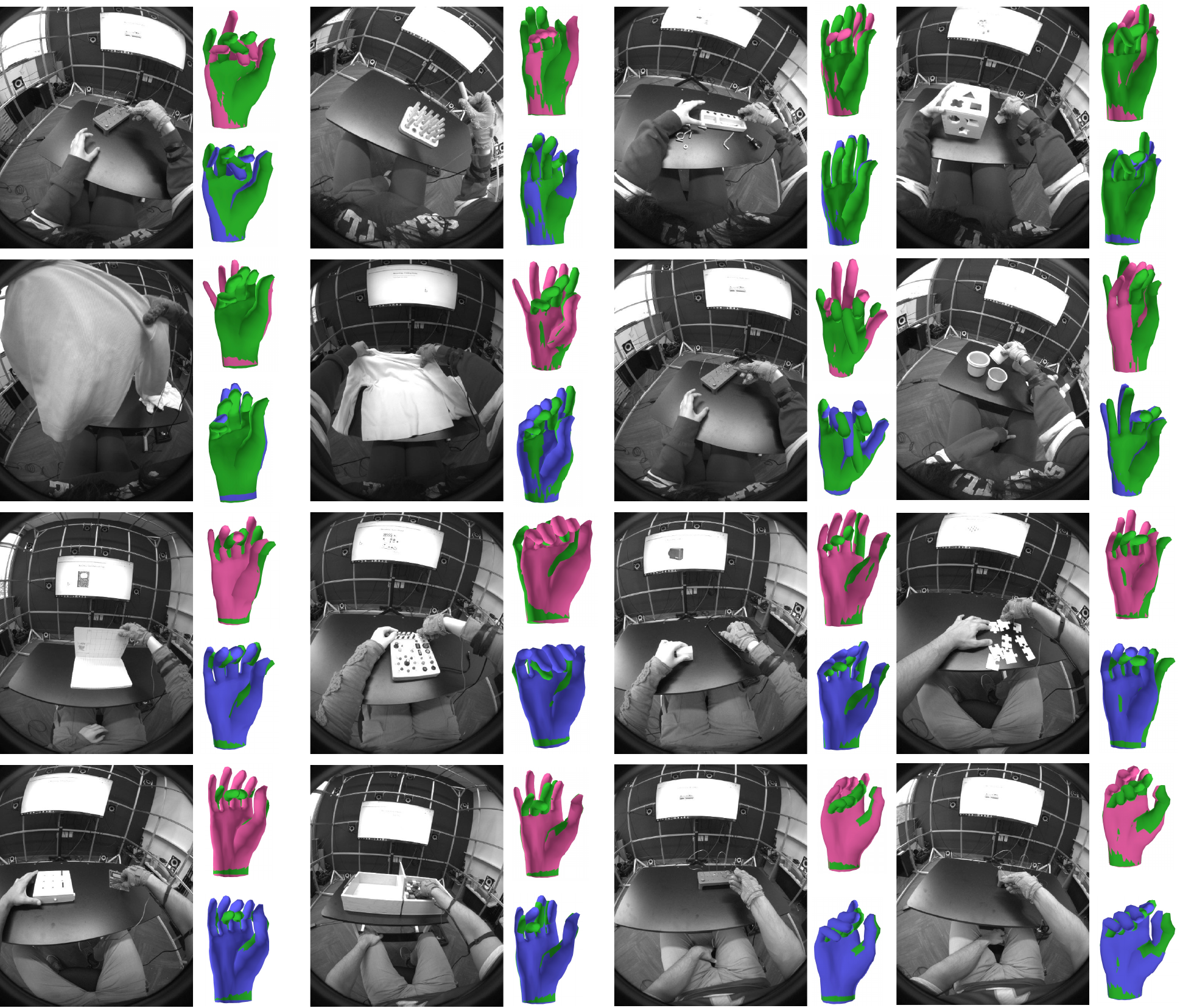}
    \caption{More qualitative comparison between AVI-HT and UMETrack.}
    \label{fig:more_vis}
\end{figure}

We provide additional qualitative comparisons between AVI-HT and UMETrack in Figure \ref{fig:more_vis}, illustrating that AVI-HT achieves more accurate hand tracking during hand-object interaction under visual occlusion. 
We also include a video in \textcolor{myblue}{supplementary.zip} that demonstrates AVI-HT's advantage in a sequential setting. As shown in the video, when the gloved hand becomes occluded, UMETrack's tracking error increases substantially, whereas AVI-HT largely maintains accurate pose estimates. In the right sub-plot, we visualize the 2nd vision-IMU attention weights: the importance assigned to IMU signals rises markedly during occlusion and recedes as hand visibility is restored. The results confirm that the model learns to dynamically leverage inertial cues when visual evidence varies.

\section{Sensitivity Study}
\label{sec:exp_sensitivity}

\begin{figure}[t]
    \centering

\includegraphics[width=0.8\linewidth]{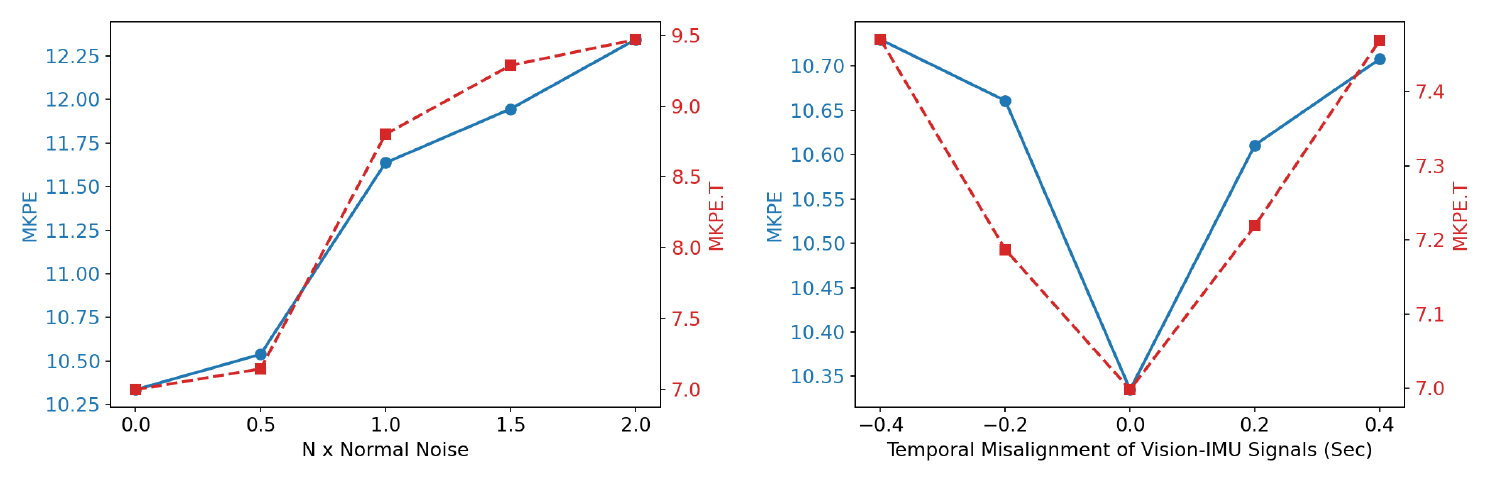}
    \caption{Sensitivity to IMU signal quality. (Left) MKPE and MKPE.T under increasing additive Gaussian noise. (Right) MKPE and MKPE.T under temporal misalignment between vision and IMU.}
    \label{fig:imu_sens}
\end{figure}

To assess the robustness of AVI-HT to degraded IMU signals, we conduct
two sensitivity analyses shown in Figure \ref{fig:imu_sens}. We firstly inject additive Gaussian noise
into the IMU input at increasing levels from $\times 0$ to $\times 2$ of
the native noise floor and train and evaluate AVI-HT in terms of MKPE and
MKPE.T metrics. We observe that AVI-HT degrades with MKPE rising from
${\sim}$10.335\,mm to ${\sim}$12.344\,mm at $2\times$
noise, while MKPE.T increases from ${\sim}$6.998\,mm to
${\sim}$9.469\,mm. The results indicate that AVI-HT is robust to moderate perturbations (e.g., $<\times 0.5$) but suffers more with higher perturbations. 
We then artificially shift the IMU window
relative to the visual frame by $-0.4$ to $+0.4$ seconds and measure
MKPE and MKPE.T for AVI-HT-UME. Both error curves form a clear V-shape centered at
zero shift, which confirms that the temporal alignment between modalities is
critical. 
We observe that the performance is relatively more stable than the noise injection, which suggests that moderate temporal misalignment preserves the overall motion dynamics within the IMU window.

\newpage
\section*{NeurIPS Paper Checklist}

\begin{enumerate}

\item {\bf Claims}
    \item[] Question: Do the main claims made in the abstract and introduction accurately reflect the paper's contributions and scope?
    \item[] Answer: \answerYes{} 
    \item[] Justification: the main claims made in the abstract and introduction accurately reflect the paper's contributions and scope.
    \item[] Guidelines:
    \begin{itemize}
        \item The answer \answerNA{} means that the abstract and introduction do not include the claims made in the paper.
        \item The abstract and/or introduction should clearly state the claims made, including the contributions made in the paper and important assumptions and limitations. A \answerNo{} or \answerNA{} answer to this question will not be perceived well by the reviewers. 
        \item The claims made should match theoretical and experimental results, and reflect how much the results can be expected to generalize to other settings. 
        \item It is fine to include aspirational goals as motivation as long as it is clear that these goals are not attained by the paper. 
    \end{itemize}

\item {\bf Limitations}
    \item[] Question: Does the paper discuss the limitations of the work performed by the authors?
    \item[] Answer: \answerYes{} 
    \item[] Justification: The paper discuss the limitations of the work performed by the authors.
    \item[] Guidelines:
    \begin{itemize}
        \item The answer \answerNA{} means that the paper has no limitation while the answer \answerNo{} means that the paper has limitations, but those are not discussed in the paper. 
        \item The authors are encouraged to create a separate ``Limitations'' section in their paper.
        \item The paper should point out any strong assumptions and how robust the results are to violations of these assumptions (e.g., independence assumptions, noiseless settings, model well-specification, asymptotic approximations only holding locally). The authors should reflect on how these assumptions might be violated in practice and what the implications would be.
        \item The authors should reflect on the scope of the claims made, e.g., if the approach was only tested on a few datasets or with a few runs. In general, empirical results often depend on implicit assumptions, which should be articulated.
        \item The authors should reflect on the factors that influence the performance of the approach. For example, a facial recognition algorithm may perform poorly when image resolution is low or images are taken in low lighting. Or a speech-to-text system might not be used reliably to provide closed captions for online lectures because it fails to handle technical jargon.
        \item The authors should discuss the computational efficiency of the proposed algorithms and how they scale with dataset size.
        \item If applicable, the authors should discuss possible limitations of their approach to address problems of privacy and fairness.
        \item While the authors might fear that complete honesty about limitations might be used by reviewers as grounds for rejection, a worse outcome might be that reviewers discover limitations that aren't acknowledged in the paper. The authors should use their best judgment and recognize that individual actions in favor of transparency play an important role in developing norms that preserve the integrity of the community. Reviewers will be specifically instructed to not penalize honesty concerning limitations.
    \end{itemize}

\item {\bf Theory assumptions and proofs}
    \item[] Question: For each theoretical result, does the paper provide the full set of assumptions and a complete (and correct) proof?
    \item[] Answer: \answerNA{} 
    \item[] Justification: This paper does not include theoretical results; our contributions are empirical.
    \item[] Guidelines:
    \begin{itemize}
        \item The answer \answerNA{} means that the paper does not include theoretical results. 
        \item All the theorems, formulas, and proofs in the paper should be numbered and cross-referenced.
        \item All assumptions should be clearly stated or referenced in the statement of any theorems.
        \item The proofs can either appear in the main paper or the supplemental material, but if they appear in the supplemental material, the authors are encouraged to provide a short proof sketch to provide intuition. 
        \item Inversely, any informal proof provided in the core of the paper should be complemented by formal proofs provided in appendix or supplemental material.
        \item Theorems and Lemmas that the proof relies upon should be properly referenced. 
    \end{itemize}

    \item {\bf Experimental result reproducibility}
    \item[] Question: Does the paper fully disclose all the information needed to reproduce the main experimental results of the paper to the extent that it affects the main claims and/or conclusions of the paper (regardless of whether the code and data are provided or not)?
    \item[] Answer: \answerYes{} 
    \item[] Justification: The paper provides complete architectural details including the IMU encoder, vision encoders, the per-joint
   cross-sensor attention mechanism with kinematic prior mask , and all training losses and objectives (Section \ref{sec:architecture}). Input preprocessing, IMU temporal windowing, and        
  synchronization procedures are described in Section \ref{input_signals}. The DexGloveHOI benchmark, including capture setup, gesture protocol, and ground-truth generation, is fully specified in Section \ref{sec:eval_data}.
    \item[] Guidelines:
    \begin{itemize}
        \item The answer \answerNA{} means that the paper does not include experiments.
        \item If the paper includes experiments, a \answerNo{} answer to this question will not be perceived well by the reviewers: Making the paper reproducible is important, regardless of whether the code and data are provided or not.
        \item If the contribution is a dataset and\slash or model, the authors should describe the steps taken to make their results reproducible or verifiable. 
        \item Depending on the contribution, reproducibility can be accomplished in various ways. For example, if the contribution is a novel architecture, describing the architecture fully might suffice, or if the contribution is a specific model and empirical evaluation, it may be necessary to either make it possible for others to replicate the model with the same dataset, or provide access to the model. In general. releasing code and data is often one good way to accomplish this, but reproducibility can also be provided via detailed instructions for how to replicate the results, access to a hosted model (e.g., in the case of a large language model), releasing of a model checkpoint, or other means that are appropriate to the research performed.
        \item While NeurIPS does not require releasing code, the conference does require all submissions to provide some reasonable avenue for reproducibility, which may depend on the nature of the contribution. For example
        \begin{enumerate}
            \item If the contribution is primarily a new algorithm, the paper should make it clear how to reproduce that algorithm.
            \item If the contribution is primarily a new model architecture, the paper should describe the architecture clearly and fully.
            \item If the contribution is a new model (e.g., a large language model), then there should either be a way to access this model for reproducing the results or a way to reproduce the model (e.g., with an open-source dataset or instructions for how to construct the dataset).
            \item We recognize that reproducibility may be tricky in some cases, in which case authors are welcome to describe the particular way they provide for reproducibility. In the case of closed-source models, it may be that access to the model is limited in some way (e.g., to registered users), but it should be possible for other researchers to have some path to reproducing or verifying the results.
        \end{enumerate}
    \end{itemize}

\item {\bf Open access to data and code}
    \item[] Question: Does the paper provide open access to the data and code, with sufficient instructions to faithfully reproduce the main experimental results, as described in supplemental material?
    \item[] Answer: \answerNo{} 
    \item[] Justification: Both the code and datasets are subject to institutional privacy and security review requirements. We are unable to provide open access at this time but are
  actively pursuing the necessary internal review processes to enable public release in the future. We provide detailed descriptions of our model architecture, training procedures, and   
  evaluation protocol in the paper and supplementary material to support reproducibility. 
    \item[] Guidelines:
    \begin{itemize}
        \item The answer \answerNA{} means that paper does not include experiments requiring code.
        \item Please see the NeurIPS code and data submission guidelines (\url{https://neurips.cc/public/guides/CodeSubmissionPolicy}) for more details.
        \item While we encourage the release of code and data, we understand that this might not be possible, so \answerNo{} is an acceptable answer. Papers cannot be rejected simply for not including code, unless this is central to the contribution (e.g., for a new open-source benchmark).
        \item The instructions should contain the exact command and environment needed to run to reproduce the results. See the NeurIPS code and data submission guidelines (\url{https://neurips.cc/public/guides/CodeSubmissionPolicy}) for more details.
        \item The authors should provide instructions on data access and preparation, including how to access the raw data, preprocessed data, intermediate data, and generated data, etc.
        \item The authors should provide scripts to reproduce all experimental results for the new proposed method and baselines. If only a subset of experiments are reproducible, they should state which ones are omitted from the script and why.
        \item At submission time, to preserve anonymity, the authors should release anonymized versions (if applicable).
        \item Providing as much information as possible in supplemental material (appended to the paper) is recommended, but including URLs to data and code is permitted.
    \end{itemize}

\item {\bf Experimental setting/details}
    \item[] Question: Does the paper specify all the training and test details (e.g., data splits, hyperparameters, how they were chosen, type of optimizer) necessary to understand the results?
    \item[] Answer: \answerYes{} 
    \item[] Justification: We specified the details of the model architecture in Section \ref{sec:architecture} and training/testing implemenetation details in the second paragraph of Section \ref{sec:experiments}..
    \item[] Guidelines:
    \begin{itemize}
        \item The answer \answerNA{} means that the paper does not include experiments.
        \item The experimental setting should be presented in the core of the paper to a level of detail that is necessary to appreciate the results and make sense of them.
        \item The full details can be provided either with the code, in appendix, or as supplemental material.
    \end{itemize}

\item {\bf Experiment statistical significance}
    \item[] Question: Does the paper report error bars suitably and correctly defined or other appropriate information about the statistical significance of the experiments?
    \item[] Answer: \answerNo{} 
    \item[] Justification: we do not report statistical confidence intervals over multiple training runs since evaluation is deterministic on a fixed test set, the main source of variability would be training initialization, which we did not sweep.
    \item[] Guidelines:
    \begin{itemize}
        \item The answer \answerNA{} means that the paper does not include experiments.
        \item The authors should answer \answerYes{} if the results are accompanied by error bars, confidence intervals, or statistical significance tests, at least for the experiments that support the main claims of the paper.
        \item The factors of variability that the error bars are capturing should be clearly stated (for example, train/test split, initialization, random drawing of some parameter, or overall run with given experimental conditions).
        \item The method for calculating the error bars should be explained (closed form formula, call to a library function, bootstrap, etc.)
        \item The assumptions made should be given (e.g., Normally distributed errors).
        \item It should be clear whether the error bar is the standard deviation or the standard error of the mean.
        \item It is OK to report 1-sigma error bars, but one should state it. The authors should preferably report a 2-sigma error bar than state that they have a 96\% CI, if the hypothesis of Normality of errors is not verified.
        \item For asymmetric distributions, the authors should be careful not to show in tables or figures symmetric error bars that would yield results that are out of range (e.g., negative error rates).
        \item If error bars are reported in tables or plots, the authors should explain in the text how they were calculated and reference the corresponding figures or tables in the text.
    \end{itemize}

\item {\bf Experiments compute resources}
    \item[] Question: For each experiment, does the paper provide sufficient information on the computer resources (type of compute workers, memory, time of execution) needed to reproduce the experiments?
    \item[] Answer: \answerYes{} 
    \item[] Justification: we provide sufficient information on the computer resources in the second paragraph of Section \ref{sec:experiments}.
    \item[] Guidelines:
    \begin{itemize}
        \item The answer \answerNA{} means that the paper does not include experiments.
        \item The paper should indicate the type of compute workers CPU or GPU, internal cluster, or cloud provider, including relevant memory and storage.
        \item The paper should provide the amount of compute required for each of the individual experimental runs as well as estimate the total compute. 
        \item The paper should disclose whether the full research project required more compute than the experiments reported in the paper (e.g., preliminary or failed experiments that didn't make it into the paper). 
    \end{itemize}
    
\item {\bf Code of ethics}
    \item[] Question: Does the research conducted in the paper conform, in every respect, with the NeurIPS Code of Ethics \url{https://neurips.cc/public/EthicsGuidelines}?
    \item[] Answer: \answerYes{} 
    \item[] Justification: the research conducted in the paper conform, in every respect, with the NeurIPS Code of Ethics.
    \item[] Guidelines:
    \begin{itemize}
        \item The answer \answerNA{} means that the authors have not reviewed the NeurIPS Code of Ethics.
        \item If the authors answer \answerNo, they should explain the special circumstances that require a deviation from the Code of Ethics.
        \item The authors should make sure to preserve anonymity (e.g., if there is a special consideration due to laws or regulations in their jurisdiction).
    \end{itemize}

\item {\bf Broader impacts}
    \item[] Question: Does the paper discuss both potential positive societal impacts and negative societal impacts of the work performed?
    \item[] Answer: \answerNA{} 
    \item[] Justification: there is no societal impact of the work performed.
    \item[] Guidelines:
    \begin{itemize}
        \item The answer \answerNA{} means that there is no societal impact of the work performed.
        \item If the authors answer \answerNA{} or \answerNo, they should explain why their work has no societal impact or why the paper does not address societal impact.
        \item Examples of negative societal impacts include potential malicious or unintended uses (e.g., disinformation, generating fake profiles, surveillance), fairness considerations (e.g., deployment of technologies that could make decisions that unfairly impact specific groups), privacy considerations, and security considerations.
        \item The conference expects that many papers will be foundational research and not tied to particular applications, let alone deployments. However, if there is a direct path to any negative applications, the authors should point it out. For example, it is legitimate to point out that an improvement in the quality of generative models could be used to generate Deepfakes for disinformation. On the other hand, it is not needed to point out that a generic algorithm for optimizing neural networks could enable people to train models that generate Deepfakes faster.
        \item The authors should consider possible harms that could arise when the technology is being used as intended and functioning correctly, harms that could arise when the technology is being used as intended but gives incorrect results, and harms following from (intentional or unintentional) misuse of the technology.
        \item If there are negative societal impacts, the authors could also discuss possible mitigation strategies (e.g., gated release of models, providing defenses in addition to attacks, mechanisms for monitoring misuse, mechanisms to monitor how a system learns from feedback over time, improving the efficiency and accessibility of ML).
    \end{itemize}
    
\item {\bf Safeguards}
    \item[] Question: Does the paper describe safeguards that have been put in place for responsible release of data or models that have a high risk for misuse (e.g., pre-trained language models, image generators, or scraped datasets)?
    \item[] Answer: \answerNA{} 
    \item[] Justification: the paper poses no such risks.
    \item[] Guidelines:
    \begin{itemize}
        \item The answer \answerNA{} means that the paper poses no such risks.
        \item Released models that have a high risk for misuse or dual-use should be released with necessary safeguards to allow for controlled use of the model, for example by requiring that users adhere to usage guidelines or restrictions to access the model or implementing safety filters. 
        \item Datasets that have been scraped from the Internet could pose safety risks. The authors should describe how they avoided releasing unsafe images.
        \item We recognize that providing effective safeguards is challenging, and many papers do not require this, but we encourage authors to take this into account and make a best faith effort.
    \end{itemize}

\item {\bf Licenses for existing assets}
    \item[] Question: Are the creators or original owners of assets (e.g., code, data, models), used in the paper, properly credited and are the license and terms of use explicitly mentioned and properly respected?
    \item[] Answer: \answerYes{} 
    \item[] Justification: the creators or original owners of assets (e.g., code, data, models), used in the paper, properly credited and are the license and terms of use explicitly mentioned and properly respectedthe creators or original owners of assets (e.g., code, data, models), used in the paper, properly credited and are the license and terms of use explicitly mentioned and properly respected.
    \item[] Guidelines:
    \begin{itemize}
        \item The answer \answerNA{} means that the paper does not use existing assets.
        \item The authors should cite the original paper that produced the code package or dataset.
        \item The authors should state which version of the asset is used and, if possible, include a URL.
        \item The name of the license (e.g., CC-BY 4.0) should be included for each asset.
        \item For scraped data from a particular source (e.g., website), the copyright and terms of service of that source should be provided.
        \item If assets are released, the license, copyright information, and terms of use in the package should be provided. For popular datasets, \url{paperswithcode.com/datasets} has curated licenses for some datasets. Their licensing guide can help determine the license of a dataset.
        \item For existing datasets that are re-packaged, both the original license and the license of the derived asset (if it has changed) should be provided.
        \item If this information is not available online, the authors are encouraged to reach out to the asset's creators.
    \end{itemize}

\item {\bf New assets}
    \item[] Question: Are new assets introduced in the paper well documented and is the documentation provided alongside the assets?
    \item[] Answer: \answerNA{} 
    \item[] Justification: the paper does not release new assets.
    \item[] Guidelines:
    \begin{itemize}
        \item The answer \answerNA{} means that the paper does not release new assets.
        \item Researchers should communicate the details of the dataset\slash code\slash model as part of their submissions via structured templates. This includes details about training, license, limitations, etc. 
        \item The paper should discuss whether and how consent was obtained from people whose asset is used.
        \item At submission time, remember to anonymize your assets (if applicable). You can either create an anonymized URL or include an anonymized zip file.
    \end{itemize}

\item {\bf Crowdsourcing and research with human subjects}
    \item[] Question: For crowdsourcing experiments and research with human subjects, does the paper include the full text of instructions given to participants and screenshots, if applicable, as well as details about compensation (if any)? 
    \item[] Answer: \answerNA{} 
    \item[] Justification: the paper does not involve crowdsourcing nor research with human subjects.
    \item[] Guidelines:
    \begin{itemize}
        \item The answer \answerNA{} means that the paper does not involve crowdsourcing nor research with human subjects.
        \item Including this information in the supplemental material is fine, but if the main contribution of the paper involves human subjects, then as much detail as possible should be included in the main paper. 
        \item According to the NeurIPS Code of Ethics, workers involved in data collection, curation, or other labor should be paid at least the minimum wage in the country of the data collector. 
    \end{itemize}

\item {\bf Institutional review board (IRB) approvals or equivalent for research with human subjects}
    \item[] Question: Does the paper describe potential risks incurred by study participants, whether such risks were disclosed to the subjects, and whether Institutional Review Board (IRB) approvals (or an equivalent approval/review based on the requirements of your country or institution) were obtained?
    \item[] Answer: \answerNA{} 
    \item[] Justification: the paper does not involve crowdsourcing nor research with human subjects.
    \item[] Guidelines:
    \begin{itemize}
        \item The answer \answerNA{} means that the paper does not involve crowdsourcing nor research with human subjects.
        \item Depending on the country in which research is conducted, IRB approval (or equivalent) may be required for any human subjects research. If you obtained IRB approval, you should clearly state this in the paper. 
        \item We recognize that the procedures for this may vary significantly between institutions and locations, and we expect authors to adhere to the NeurIPS Code of Ethics and the guidelines for their institution. 
        \item For initial submissions, do not include any information that would break anonymity (if applicable), such as the institution conducting the review.
    \end{itemize}

\item {\bf Declaration of LLM usage}
    \item[] Question: Does the paper describe the usage of LLMs if it is an important, original, or non-standard component of the core methods in this research? Note that if the LLM is used only for writing, editing, or formatting purposes and does \emph{not} impact the core methodology, scientific rigor, or originality of the research, declaration is not required.
    \item[] Answer: \answerNA{} 
    \item[] Justification: the core method development in this research does not involve LLMs as any important, original, or non-standard components.
    \item[] Guidelines:
    \begin{itemize}
        \item The answer \answerNA{} means that the core method development in this research does not involve LLMs as any important, original, or non-standard components.
        \item Please refer to our LLM policy in the NeurIPS handbook for what should or should not be described.
    \end{itemize}

\end{enumerate}
\end{document}